\documentclass[journal]{formats/IEEEtran}

\usepackage[backend=bibtex,
            hyperref=false,
            url=false,
            isbn=false,
            doi=false,
            backref=false,
            style=ieee,
            citestyle=numeric-comp,
            sorting=nyt,
            block=none]{biblatex}

\usepackage{flushend}
\usepackage{balance}
\usepackage{algorithm}
\usepackage{algpseudocode}

\usepackage{float}
\usepackage{placeins}
\usepackage{graphics}
\usepackage{algorithm}
\usepackage{algpseudocode}
\usepackage[pdftex]{graphicx}
\DeclareGraphicsExtensions{.pdf,.png,.jpg}

\usepackage[skip=2pt,font=small]{caption}
\usepackage{subcaption}
\usepackage[rightcaption]{sidecap}
\usepackage{pbox}

\usepackage{mathtools}
\usepackage{amsmath, amssymb, amscd}
\usepackage{ wasysym } 
\usepackage{amsfonts}
\usepackage{mathptmx} 
\usepackage{gensymb} 
\usepackage{nicefrac}       

\DeclareMathAlphabet{\mathcal}{OMS}{lmsy}{m}{n}
\DeclareSymbolFont{largesymbols}{OMX}{cmex}{m}{n}
\usepackage{textcomp} 

\usepackage{algorithm} 
\usepackage{algorithmicx, algpseudocode} 

\usepackage{array} 
\usepackage{multirow}
\usepackage{multicol}

\usepackage[T1]{fontenc} 
\usepackage[utf8]{inputenc}
\usepackage[english]{babel} 
\usepackage{units}
\usepackage{bm}
\usepackage{times} 
\usepackage{xspace}
\usepackage{flushend}
\usepackage{csquotes}
\usepackage{makeidx}
\usepackage{blindtext}
\let\labelindent\relax
\usepackage{enumitem}

\usepackage{soul} 
\usepackage{subfiles} 

\usepackage[yyyymmdd]{datetime}

\date{\protect\formatdate{1}{1}{2001}}

\usepackage{url}
\makeatletter
\g@addto@macro{\UrlBreaks}{\UrlOrds}
\makeatother
\usepackage{color}
\usepackage[usenames,dvipsnames,table]{xcolor}

\usepackage{marginnote}
\usepackage{soul} 

\usepackage[colorinlistoftodos]{todonotes}

\newcommand{\jean}[1]{\todo[inline,color=red!40]{J: #1}}

\newcommand{\ignore}[1]{}

\newcommand{\figref}[1]{Fig.    \,\ref{fig:#1}}

\newcommand{\secref}[1]{Sec.\,\ref{sec:#1}}

\newcommand{\dataset}{\mathcal{D}}
\newcommand{\iobs}{\mathbf{o}_i}
\newcommand{\ilabels}{\mathbf{y}_i}
\newcommand{\latent}{\mathbf{z}}

\newcommand{\piparams}{\mathbf{\theta}_\pi}
\newcommand{\repparams}{\mathbf{\theta}_s}

\newcommand{\markov}{\mathcal{M}}

\newcommand{\tstate}{\mathbf{s}_t}
\newcommand{\states}{\mathcal{S}}
\newcommand{\actions}{\mathcal{A}}
\newcommand{\action}{\mathbf{a}}
\newcommand{\taction}{\mathbf{a}_t}
\newcommand{\stateinitial}{\rho_0}
\newcommand{\rewardfn}{r}
\newcommand{\transition}{\mathcal{T}}
\newcommand{\timemax}{T}
\newcommand{\policy}{\pi}

\newcommand{\ryinit}{\beta_\textrm{init}}
\newcommand{\rxinit}{\gamma\textrm{init}}
\newcommand{\rz}{\alpha}
\newcommand{\rx}{\gamma}
\newcommand{\ry}{\beta}
\newcommand{\rzdelta}{\Delta \alpha}

\newcommand{\poset}{\mathbf{p_t}}
\newcommand{\posedes}{\mathbf{p}_{\textrm{des}}}

\newcommand{\posedelta}{\Delta \mathbf{p}}
\newcommand{\cartdelta}{\Delta \mathbf{x}}
\newcommand{\cartpos}{\mathbf{x}}

\newcommand{\cartposd}{\mathbf{x}_{\textrm{des}}}

\newcommand{\cartvel}{\mathbf{v}}
\newcommand{\cartveld}{\mathbf{v}_{\textrm{des}}}
\newcommand{\cartveldt}{\mathbf{v}_k}

\newcommand{\cartaccd}{\mathbf{a}_{\textrm{des}}}
\newcommand{\cartaccdt}{\mathbf{a}_k}
\newcommand{\cartaccu}{\mathbf{a}_u}

\newcommand{\kp}{\mathbf{k}_{\textrm{p}}}
\newcommand{\kv}{\mathbf{k}_{\textrm{v}}}
\newcommand{\taurobot}{\mathbf{\tau}_{\textrm{u}}}
\newcommand{\varpoe}{\mathbf{\sigma}^2}
\newcommand{\mupoe}{\mathbf{\mu}}

\makeatletter
\newenvironment{mcases}[1][l]
 {\let\@ifnextchar\new@ifnextchar
  \left\lbrace
  \array{@{}l@{\quad}#1@{}}}
 {\endarray\right.}
\makeatother

\begin{document}


%


\title{Making Sense of Vision and Touch: Learning Multimodal Representations for Contact-Rich Tasks}

\author{Michelle A. Lee, Yuke Zhu, Peter Zachares, Matthew Tan, Krishnan Srinivasan, Silvio Savarese, Li Fei-Fei, Animesh Garg, Jeannette Bohg \thanks{\vspace{-10pt} \hrule \vspace{1pt}Authors are with the Department of Computer Science, Stanford University. {\tt [mishlee,yukez,zachares,mratan,krshna,ssilvio,feifeili, animeshg,bohg]@stanford.edu}. A. Garg is also at Nvidia, USA.}}


\maketitle

\begin{abstract}
Contact-rich manipulation tasks in unstructured environments often require both haptic and visual feedback. It is non-trivial to manually design a robot controller that combines these modalities which have very different characteristics. While deep reinforcement learning has shown success in learning control policies for high-dimensional inputs, these algorithms are generally intractable to deploy on real robots due to sample complexity. In this work, we use self-supervision to learn a compact and multimodal representation of our sensory inputs, which can then be used to improve the sample efficiency of our policy learning. Evaluating our method on a peg insertion task, we show that it generalizes over varying geometries, configurations, and clearances, while being robust to external perturbations. We also systematically study different self-supervised learning objectives and representation learning architectures. Results are presented in simulation and on a physical robot.  


\end{abstract}

\begin{IEEEkeywords}
Deep Learning in Robotics and Automation, Perception for Grasping and Manipulation, Sensor Fusion, Sensor-based Control 
\end{IEEEkeywords}

\IEEEpeerreviewmaketitle

\section{introduction}

\begin{figure}[t!]
\centering
\includegraphics[width=\columnwidth]{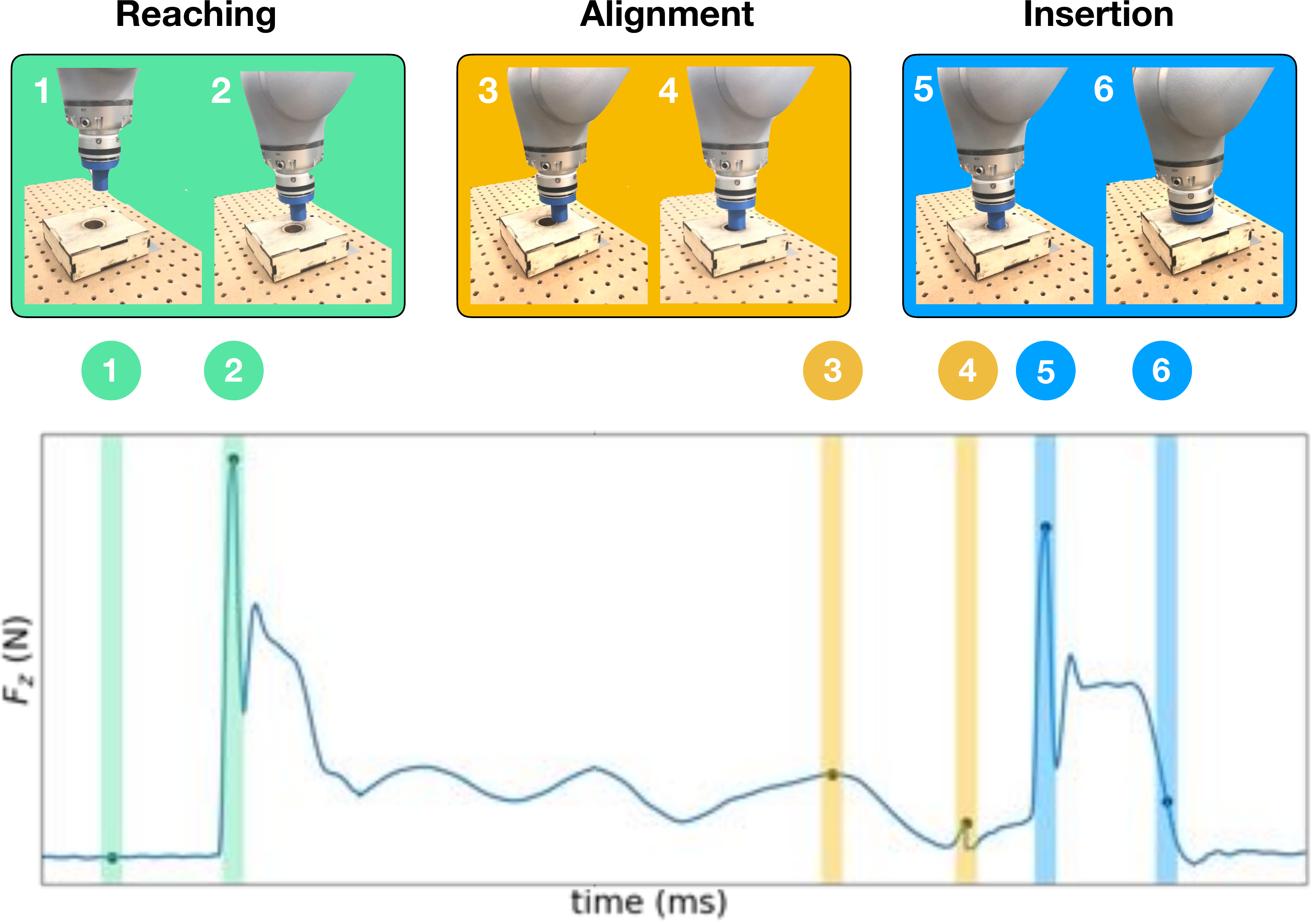}
\caption{Force sensor readings in the z-axis (height) and visual observations are shown with corresponding stages of a peg insertion task. The force reading transitions from (1) the arm moving in free space to (2) making contact with the box. While aligning the peg, the forces capture the sliding contact dynamics on the box surface (3, 4). Finally, in the insertion stage, the forces peak as the robot attempts to insert the peg at the edge of the hole (5), and decrease when the peg slides into the hole (6).}
\label{fig:insertion}
\end{figure}

\IEEEPARstart{E}{ven} in routine tasks such as inserting a car key into the ignition, humans seamlessly combine the senses of vision and touch to complete the task. Visual feedback provides semantic and geometric object properties for accurate reaching or grasp pre-shaping. Haptic feedback provides observations of current contact conditions between object and environment for accurate localization and control under occlusions. These two types of feedback modalities are complementary and concurrent during contact-rich manipulation~\cite{blake2004neural}, which is illustrated in Fig. \ref{fig:insertion}. Yet few algorithms endow robots with a similar capability. While the utility of multimodal data has frequently been shown in robotics~\cite{bicchi1988integrated,romano2011human,veiga2015stabilizing,Song:2014, fazeli2019see}, the proposed manipulation strategies often rely on handcrafted features or prior knowledge about how to solve a task. This makes many of these methods task-specific. On the other hand, most learning-based methods do not require manual task specification, yet the majority of learned manipulation policies close the control loop around a single modality, often RGB images \cite{Levine:Finn:2016,chebotar2017path, finn2017deep, zhu2018reinforcement}.

In this work, we equip a robot with a policy that leverages multimodal feedback from vision and touch -- modalities that differ in many characteristics such as dimensionality, sampling frequency, and value range. Our proposed policy is learned through self-supervision and generalizes over variations of the same contact-rich manipulation task in geometries, configurations, and clearances. As a case study, we use the task of peg insertion. We qualitatively show that the learned policy is also robust to some external perturbations. 
Our approach starts with using neural networks to learn a joint representation of haptic, RGB-D as well as proprioceptive information. Using self-supervised learning objectives, this network is trained to predict optical flow, whether contact will be made in the next control cycle, future end-effector position, and concurrency of visual and haptic data. The training is action-conditional, encouraging the encoding of action-related information. The resulting compact representation of the high-dimensional and heterogeneous data forms the input to a policy for contact-rich manipulation tasks (in this paper peg insertion) that is learned using deep reinforcement learning. The proposed decoupling of state estimation and control achieves practical sample efficiency for learning both representation and policy on a real robot. 

Our primary contributions are:
\begin{enumerate}
    \item  A variational model for multimodal representation learning from which a contact-rich manipulation policy can be learned. 
    \item Demonstration of peg insertion tasks that effectively utilize both haptic and visual feedback for hole search, peg alignment, and insertion (see Fig \ref{fig:insertion}). Ablation studies comparing the effects of each modality on task performance. 
    \item Evaluation of generalization to tasks with different peg geometry and of robustness to perturbation and sensor noise. 
\end{enumerate}

This work is an extended version of a previously published conference paper \cite{lee2018making}. We propose a new variational representation learning technique and significantly expand the experimental evaluation of the overall methodology in the following ways:

\begin{enumerate}
    \setcounter{enumi}{3}
    \item Analysis of our multimodal representation model compared to baseline models with different learning objectives, architecture, and dimension of the representation.
    \item Addition of depth as input modality, and addition of end effector roll to action space which makes the peg insertion task more challenging and increases the dimensionality of the action space from 3-DoF to 4-DoF. 
    \item Reproduction of results on a new robot, the Franka Panda.
\end{enumerate}

\section{Related Work and Background}
\subsection{Contact-Rich Manipulation}
Contact-rich tasks, such as peg insertion, fastening screws, and edge following, have been studied for decades due to their relevance in manufacturing. Manipulation policies often rely entirely on haptic feedback and force control, and assume sufficiently accurate state estimation~\cite{Whitney:1987}. They typically generalize over certain task variations, for instance, peg-in-chamfered-hole insertion policies that work independently of peg diameter~\cite{Whitney1982}. However, entirely new policies are required for new geometries. For chamferless holes, manually defining a small set of viable contact configurations has been successful~\cite{Caine99} but cannot accommodate the vast range of real-world variations. \cite{Song:2014} combine visual and haptic data for inserting two planar pegs with more complex cross sections, but assume known peg geometry.

Reinforcement learning approaches have recently been proposed to address variations in geometry and configuration for manipulation.
\cite{Levine:Finn:2016,zhu2018reinforcement} train neural network policies using RGB images and proprioceptive feedback. Their approach works well in a wide range of tasks, but the large object clearances compared to manufacturing tasks may explain the sufficiency of RGB data. A series of learning-based approaches have relied on haptic feedback for manipulation. Many of them are concerned with estimating the stability of a grasp before lifting an object~\cite{calandra2017feeling,Yasemin:2013}, even suggesting a regrasp~\cite{Yevgen:Karol:2015}. Only a few approaches learn entire manipulation policies through reinforcement only given haptic feedback~\cite{Mrinal:2011, sung2017learning,van2016stable, van2015learning, sutanto2018learning, tian2019manipulation}. While \cite{Mrinal:2011} relies on raw force-torque feedback, \cite{sung2017learning,van2016stable, sutanto2018learning} learn a low-dimensional representation of high-dimensional tactile data before learning a policy, and \cite{tian2019manipulation} learns a dynamics model of the tactile feedback in a latent space. 

Even fewer approaches exploit the complementary nature of vision and touch. Some of them extend their previous work on grasp stability estimation~\cite{YaseminRenaud,Calandra:2018}. Others perform full manipulation tasks based on multiple input modalities~\cite{Kappler-RSS-15, fazeli2019see, abu2015adaptation} but require a pre-specified manipulation graph~\cite{Kappler-RSS-15}, demonstrate only on one task ~\cite{Kappler-RSS-15, fazeli2019see}, or require human demonstration and object CAD models~\cite{abu2015adaptation}. There have been promising works that train manipulation policies in simulation and transfer them to a real robot~\cite{andrychowicz2018learning, peng2018sim, bousmalis2018using}. However, only few works focused on contact-rich tasks~\cite{fu2016one} and none relied on haptic feedback in simulation, most likely because of the lack of fidelity of contact simulation and collision modeling for articulated rigid-body systems~\cite{shameekcollisions,fazeli2017fundamental}. 

\subsection{Representation Learning for Policy Learning}
\label{sec:related:repr}
The aim of representation learning is to discover a low-dimensional feature representation of high-dimensional data that captures the information that is relevant for a particular task. In the context of reinforcement learning (which we go into more detail in Sec. \ref{sec:ps}), a good representation encodes the essential information of the state for the agent to choose its next action for a given task \cite{StateReprLearning}. A compact and low-dimensional state representation can make reinforcement learning more data efficient. 

A popular representation learning objective is reconstruction of the raw sensory input through variational autoencoders ~\cite{de2018integrating,StateReprLearning,van2016stable, yang2017deep}, which we consider as a baseline in this work.  This unsupervised objective benefits learning stability and speed, but it is also data intensive and prone to overfitting~\cite{de2018integrating}. 
When learning for control, action-conditional predictive representations can encourage the state representations to capture action-relevant information~\cite{StateReprLearning}. There are studies that attempt to predict full images when pushing objects with benign success~\cite{agrawal2016learning, babaeizadeh2017stochastic, oh2015action}. In these cases either the underlying dynamics is deterministic~\cite{oh2015action}, or the control runs at a low frequency~\cite{finn2017deep}. In contrast, we operate with haptic feedback at 1\unit{kHz} and send Cartesian control commands at 20\unit{Hz}. We use an action-conditional surrogate objective for predicting optical flow, end-effector poses, and contact events with self-supervision.

For a detailed survey of different loss functions, model architectures and training methods for representation learning, we refer to \cite{de2018integrating,StateReprLearning}. 

\begin{figure*}[t!]
\centering
\includegraphics[trim={1cm 0cm 1cm 0.7cm},clip,width=\linewidth]{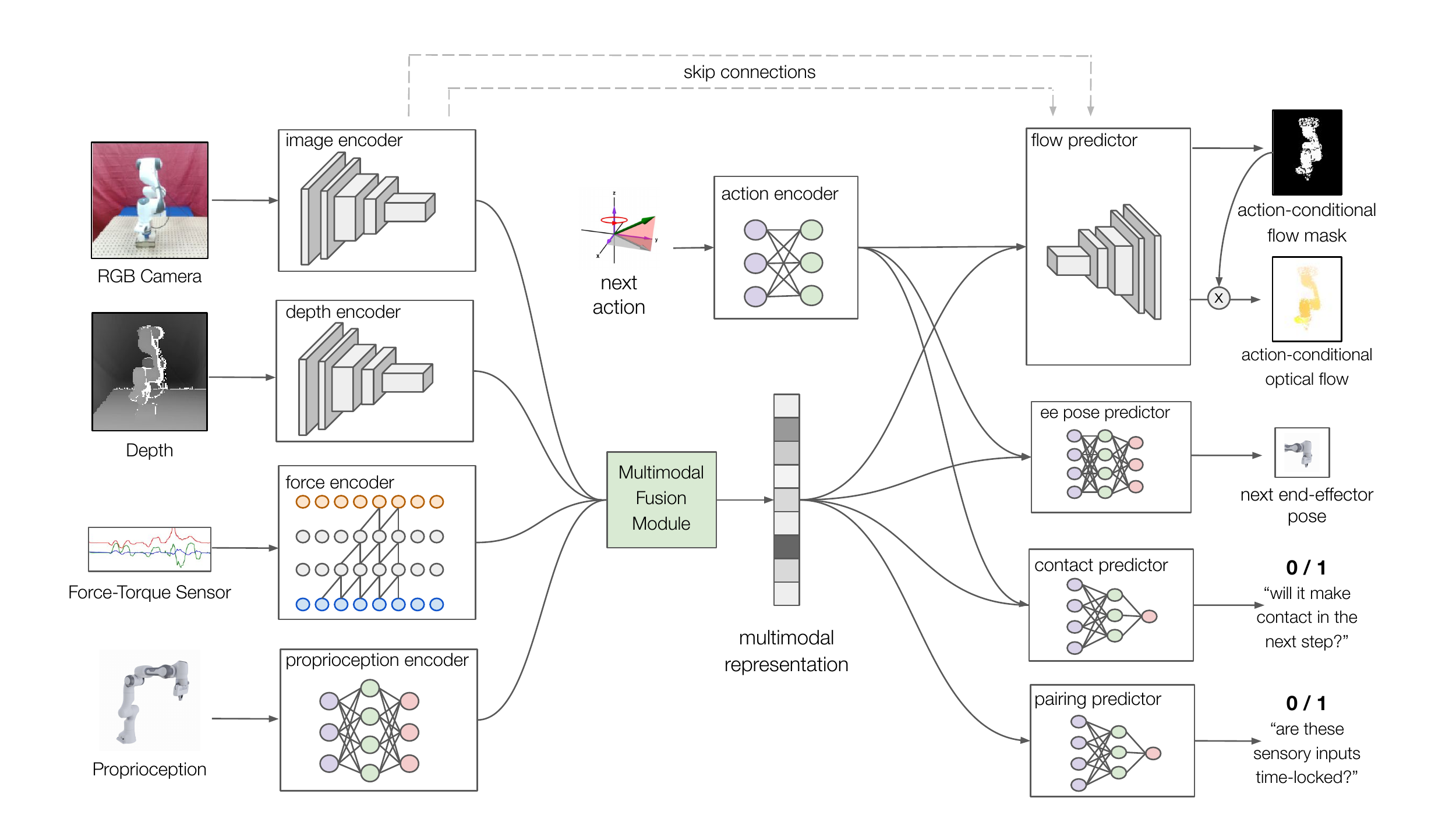}
\caption{Neural network architecture for multimodal representation learning with self-supervision. The network takes data from four different sensors as input: RGB images, depth map, F/T readings over a 32\unit{ms} window, and end-effector position, orientation, and velocity. It encodes and fuses this data into a multimodal representation using a variational Bayesian method, on which a policy for contact-rich manipulation can be learned. This representation learning network is trained end-to-end through self-supervision.}
\label{fig:network_architecture}
\end{figure*}

\subsection{Multimodal Learning}\label{sec:related:multimodal}
The complementary nature of heterogeneous sensor modalities has previously been explored for inference and decision making. In this section, we review works that include such diverse modalities as vision, range, audio, haptic and proprioceptive data as well as language. This heterogeneous data makes the application of hand-designed features and sensor fusion extremely challenging. That is why learning-based methods have been on the forefront. For example, there have been many works that have explored the correlation between auditory and visual data for tasks such as speech or material recognition or for sound source localization \cite{ngiam2011multimodal,owens2018audio,OwensIMTAF16,yang2017deep}.
\cite{Calandra:2018,gao2016deep,YaseminRenaud,SinapovSS14} fuse visual and haptic data for grasp stability assessment, manipulation, material recognition, or object categorization. \cite{liu2017learning,sung2017learning} fuse vision and range sensing and \cite{sung2017learning} add language labels. While many of these multimodal approaches are trained through a classification objective~\cite{Calandra:2018,gao2016deep,YaseminRenaud,yang2017deep}, in this paper we are interested in multimodal representation learning for control. 

There is compelling evidence that the interdependence and concurrency of different sensory streams aid perception and manipulation \cite{edelman1987neural,lacey2016crossmodal,2016_TRO_IP}. Several works have combined visual and tactile information for state estimation with probabilistic inference models such as recursive Bayesian filters \cite{martin2017cross} and factor graphs \cite{lambert2019joint, yu2018realtime}, but these methods require pre-defined visual features from visual motion trackers or patterned objects, as well as some prior knowledge of the manipulated objects, such as geometric constraints. 

In contrast, few studies have explicitly exploited this concurrency in representation learning. Examples include \cite{srivastava2012multimodal} for visual prediction tasks and \cite{ngiam2011multimodal,owens2018audio} for audio-visual coupling. In this paper, we follow \cite{WuandGoodman} who demonstrated the advantages of combining multiple, concurrent modalities into a latent space by a product-of-experts approach. Through this approach, fewer parameters and data is needed for multimodal image analysis and language translation tasks. Similar to \cite{owens2018audio}, we also adopt a self-supervised objective to fuse visual and haptic data by predicting whether visual and haptic data are temporally aligned. 

\section{Problem Statement and Method Overview}
\label{sec:ps}
Our goal is to learn a policy on a robot for performing contact-rich manipulation tasks. We want to evaluate the value of combining multisensory information and the ability to transfer multimodal representations across tasks. We systematically study different representation learning techniques and losses. 

For sample efficiency, we first learn a neural network-based feature representation of the multisensory data. The resulting compact feature vector serves as input to a neural network policy trained through deep reinforcement learning.

We model the manipulation task as a finite-horizon, discounted Markov Decision Process (MDP) $\markov$, with a state space $\states$, an action space $\actions$, state transition dynamics $\transition : \states \times \actions \to \states$, an initial state distribution $\stateinitial$, a reward function $\rewardfn : \states \times \actions \to \mathbb{R}$, horizon $\timemax$, and discount factor $\gamma_d \in (0, 1]$.
To determine the optimal stochastic policy $\policy : \states \to \mathbb{P}(\actions)$, we want to maximize the expected discounted reward 
\begin{equation}
J(\policy) = \mathbb{E}_\policy \left[\sum^{\timemax-1}_{t=0} \gamma_d^{~t}
\rewardfn(\tstate, \taction) \right]
\label{eqn:loss}
\end{equation}
We parameterize the policy with a neural network $\piparams$ that are learned as described in Sec.~\ref{sec:policy-control}. $\states$ is defined by the low-dimensional latent space representation learned from high-dimensional 2D and 3D visual data and from haptic data. This representation is a neural network parameterized by  $\phi_s$ and $\repparams$ and is trained as described in Sec.~\ref{sec:representation-learning}. We refer to this learned latent representation as $\mathbf{z}_t$ in the rest of the paper. $\actions$ is defined over continuously-valued, 3D position displacements $\cartdelta$ and roll angle displacement $\rzdelta$ in end-effector space. The controller design is detailed in Sec.~\ref{sec:policy-control}.

\section{Multi-Modal Representation Model}
\label{sec:representation-learning}
Deep networks are powerful tools to learn representations from high-dimensional data~\cite{lecun2015deep} but require a substantial amount of training data. Here, we address the challenge of seeking sources of supervision that do not rely on laborious human annotation. We design a set of predictive tasks that are suitable for learning a fused representation of visual and haptic data for contact-rich manipulation tasks, where supervision can be obtained via automatic procedures rather than manual labeling. We extend our previous work~\cite{lee2018making} by using a variational model
 for representation learning instead of a deterministic model. We show how this yields significantly better manipulation policies. Figure~\ref{fig:network_architecture} visualizes the proposed representation learning model, which uses neural network encoders to learn features from raw sensory inputs and neural network decoders to predict our self-supervised objectives. 

\subsection{Variational Inference for Representation Learning}
\label{sec:variational}
We view representation learning from the perspective of a probabilistic graphical model, where we aim to learn  $p(\latent|\dataset)$, the posterior distribution of the latent variable $\latent$ given the dataset $\dataset = \{ (\iobs, \ilabels, \action_i) | i=1\dots N\}$, which has input sensor readings $\iobs$,  self-supervised labels $\ilabels$, and robot actions $\action_i$. 

Unfortunately, computing the true posterior $p( \latent|\dataset)$ is intractable as it would require to marginalize over all possible $\latent$ for computing the evidence $p(\dataset)$. Instead, we use Variational Bayes~\cite{kingma2013auto} to find an approximate posterior $q(\latent | \dataset)$ that is as close as possible to the true posterior $p( \latent|\dataset)$. We can measure how closely $q(\latent | \dataset)$ approximates $p( \latent|\dataset)$ with the {\em Kullback-Leibler} (KL) divergence: $\mathrm{KL}(q(\latent | \dataset) || p(\latent | \dataset))$. This still requires to compute the intractable marginal likelihood $p(\dataset)$. We can rewrite the evidence as:
\begin{equation}\label{eq:loglikeD}
\begin{split}
\log p(\dataset) &= \mathrm{KL}(q(\latent | \dataset) || p(\latent | \dataset)) + \mathcal{L}(\repparams, \phi_s)\\
\end{split}
\end{equation}
where the KL divergence term is always positive from Jensen's inequality. Therefore, the second term $\mathcal{L}(\repparams, \phi_s)$ forms the {\em Evidence Lower BOund\/} (ELBO) and maximizing this bound leads to minimizing the KL divergence term.

We assume that each data point $\dataset_i= (\iobs, \ilabels, \action_i)$ maps to a unique $\latent$ such that the ELBO is a sum over a term per data point defined as:
\begin{align}
  \mathcal{L}_i(\repparams, \phi_s) &=   \mathbb{E}_{q_{\phi_s}(\latent | \dataset_i))}[\log p_{\theta_s}(\dataset_i | \latent)] \label{eqn:likelihood}  \\
  &\ - \mathrm{KL}(q_{\phi_s}(\latent | \dataset_i)||p(\latent)) \label{eqn:kl}
\end{align}
The first term (Eqn. \ref{eqn:likelihood}) refers to the expected log-likelihood of the $i$-th data point $\dataset_i$ given the latent variable $\latent$. We model the likelihood $p_{\theta_s}(\dataset_i | \latent)$ with a decoder neural network, parameterized by $\theta_s$. The second term (Eqn. \ref{eqn:kl}) can be seen as a regularization term, where we fit the posterior estimate $q_{\phi_s}(\latent | \dataset_i)$ to a prior $p(\latent)$ by minimizing the KL-divergence between the two distributions.  We model $q_{\phi_s}(\latent | \dataset_i)$ as a neural network encoder parameterized by $\phi_s$. We assume that the prior $p(\latent)$ has a standard isotropic multivariate Gaussian distribution $N(0,\mathbb{I})$. This prior has the effect that posterior estimates which diverge from a standard normal distribution incur a large penalty. To learn the parameters of the encoder and decoder networks, we minimize the negative ELBO $-\mathcal{L}(\repparams, \phi_s)=-\sum^{|\dataset|}_{i=1}\mathcal{L}_i(\repparams, \phi_s)$ with respect to $\theta_s$ and $\phi_s$. This is equivalent to maximizing Eqn~\ref{eq:loglikeD}, the log-likelihood of the data, with respect to the network parameters.

Our variational inference framework is similar to the widely used Variational Autoencoders (VAE)~\cite{kingma2013auto} with one key difference: our decoders do not reconstruct the input data. Instead, the decoders are optimizing self-supervised learning objectives such as making action-conditional predictions over time. Details of our predictive tasks and and loss functions can be found in Sec.~\ref{sec:loss}. In Sec.~\ref{sec:exp}, we will show that these predictive objectives lead to better performances than reconstruction as a representation learning objective.



\subsection{Modality Encoders}

Our model encodes four types of sensory data available to the robot: RGB ($\mathbf{o}_{\scriptscriptstyle{RGB}}$)
and depth images ($\mathbf{o}_{\scriptscriptstyle{depth}}$) from a fixed RGB-D camera, haptic feedback from a wrist-mounted force-torque (F/T) sensor ($\mathbf{o}_{\scriptscriptstyle{force}}$) , and proprioceptive data from the joint encoders of the robot arm ($\mathbf{o}_{\scriptscriptstyle{prop}}$). The heterogeneous nature of this data requires domain-specific encoders to capture the unique characteristics of each modality, which we will fuse into a single latent representation vector $\mathbf{z}$ of dimension $d$. As we are using a variational inference approach that fits the encoder's posterior distribution to an isotropic multivariate Gaussian prior, each encoder outputs a mean $\boldsymbol{\mu}$ and variance $\boldsymbol{\sigma^2}$ for each of dimension $d$. 

For visual feedback, we use a six-layer {\em convolutional neural network\/} (CNN) similar to FlowNet~\cite{flownet1} to encode $128\times 128\times 3$ RGB images. For depth feedback, we use an eighteen-layer CNN with $3\times 3$ convolutional filters of increasing depths similar to VGG-16~\cite{simonyan2014very} to encode $128\times128\times1$ depth images.  We add a single fully-connected layer to the end of both the depth and RGB encoders to transform the final activation maps into a $2\times d$-dimensional variational parameter vector. For haptic feedback, we take the last 32 readings from the six-axis F/T sensor as a $32\times 6$ time series and perform 5-layer causal convolutions~\cite{oord2016wavenet} with stride 2 to transform the force readings into a $2\times d$-dimensional variational parameter vector. For proprioception, we encode the current position, roll, linear velocity and roll angular velocity of the end-effector with a 4-layer {\em multilayer perceptron\/} (MLP) to produce a $2\times d$-dimensional variational parameter vector. In the next section, we discuss how the resulting four vectors that represent each modality are fused into one vector containing the $2\times d$-dimensional variational parameters for the latent space distribution. As default, we set $d$ to equal 128 dimensions. We analyze the sensitivity of our method to the dimensionality of the latent space representation in Sec. \ref{sec:size}. 

\subsection{Multimodal Fusion}

Following \cite{WuandGoodman}, we combine the estimated distributions of each modality using product of experts. We assume that each modality is conditionally independent given the latent variable representation $\latent$ and that each encoder maps to a multivariate isotropic Gaussian. With these assumptions, we can combine the modality-specific distributions by taking the normalized product of each Gaussian probability density function. The resulting multivariate Gaussian distribution of the multimodal latent space will have mean $\boldsymbol{\mu}$ and variance $\boldsymbol{\sigma^2}$ computed as: 

\begin{equation} \label{eq:mixture_eq}
\varpoe_{j} = \bigg(\sum_{i=1}^{n+1} \varpoe_{ij}\bigg)^{-1} \hspace{1cm} \mupoe_{j} = \bigg(\sum_{i=1}^{n+1} \mupoe_{ij}/\varpoe_{ij}\bigg)\bigg(\sum_{i=1}^{n+1} \varpoe_{ij}\bigg)^{-1} 
\end{equation}
where $n$ is the number of modalities, $\mu_j$  and $\sigma_j^2$ are the variational parameters of the $j$-th dimension of the encoder's posterior distribution, $\sigma^2_{ij}$ is the variance and $\mu_{ij}$ is the mean of the $j$-th dimension of the posterior distribution of the $i$-th modality. 
When training the representation model, we sample from the distribution represented by these variational parameters.

\subsection{Self-Supervised Predictions and Decoder Architecture} 
\label{sec:loss}

Representations that encode dynamics and action-related information have been shown to work well for policy learning \cite{StateReprLearning}. To achieve this, we design four action-conditional representation learning objectives. Given the next robot action and the compact representation of the current sensory data, the model has to predict (i) the optical flow in the image sequence generated by the action, (ii) the optical flow mask which is also used in (i), (iii) whether the end-effector will make contact with the environment in the next control cycle, and (iv) the future end-effector position. Ground-truth optical flow annotations are automatically generated given proprioception and known robot kinematics and geometry~\cite{flownet1,GarciaCifuentes.RAL}. From the ground-truth optical flow annotations we also extract the optical flow mask, which can be seen as the segmentation mask of the robot in motion. Ground-truth annotations of binary contact states are generated by applying simple heuristics which check whether the F/T readings on the wrist sensor are above certain empirically determined thresholds.

The next action, i.e. the end-effector motion, is encoded by a 2-layer MLP. The output of the action encoder is concatenated with the multimodal representation and processed by an additional 2-layer MLP, which is used as the input to the decoders. As an additional source of self-supervision not included in \cite{lee2018making}, we are also predicting action-conditional end-effector positions, which can be non-trivial to model due to errors in our dynamics model, our spline-based trajectory generator (discussed in Sec.~\ref{sec:policy-control}), and non-linear contact dynamics. 

The flow predictor uses a 4-layer convolutional decoder with upsampling to process the action-conditional feature vector. Following~\cite{flownet1}, we use 4 skip connections. At the end of these 4 layers, one convolution layer predicts the unmasked optical flow of the scene and another convolution layer predicts the optical flow mask. These two estimates are multiplied element-wise to predict the optical flow of the robot. The predicted optical flow is a $32\times32\times2$ image which is then upsampled to the size of the ground-truth optical flow $128\times128\times2$. The contact predictor is a 1-layer MLP and performs binary classification. The end-effector prediction network is a 4-layer MLP. 

As discussed in Sec.~\ref{sec:related:repr}, there is concurrency between the different sensory streams leading to correlations and redundancy, e.g., seeing the peg, touching the box, and feeling the force. We exploit this by introducing a fifth representation learning objective that predicts whether two sensor streams are temporally aligned~\cite{owens2018audio}. During training, we sample a mix of time-aligned multimodal data and randomly shifted ones which have the opposite contact state of time aligned data. The alignment predictor (a 1-layer MLP) takes the low-dimensional multimodal representation as input and performs binary classification of whether the input was aligned or not.

\subsection{Loss Functions and Training Details}

For binary classification prediction tasks, we model the likelihood distribution as a Bernoulli distribution. This allows us to use a cross entropy loss to minimize the negative log-likelihood (Eqn. \ref{eqn:likelihood}) in the negative ELBO $-\mathcal{L}(\repparams, \phi)$. For predictions with continuous values, we model the likelihood distribution with multivariate Gaussians, and use mean squared error loss functions.  

We train the action-conditional optical flow with endpoint error (EPE) loss averaged over all pixels~\cite{flownet1}, end-effector position prediction with mean squared error loss, and the contact prediction, the alignment prediction, and optical flow mask prediction with cross-entropy loss. Along with minimizing the negative log-likelihood with these five losses (Eqn. \ref{eqn:likelihood}), we also want to minimize the KL divergence (Eqn. \ref{eqn:kl}) between the approximate posterior and the prior. This gives six loss terms.



During training, we minimize the sum of the six losses described above end-to-end with stochastic gradient descent on a dataset of rolled-out trajectories. In order to backpropagate through the random variables in the proposed probabilistic network, we employ the reparametrization trick commonly used with variational inference methods described in \cite{kingma2013auto}. Once trained, this network produces a $d$-dimensional feature vector that compactly represents multimodal data. This vector is used as the input to the manipulation policy learned via reinforcement learning. Our data collection procedure is described in more detail in Sec.~\ref{sec:exp}.

\begin{figure}[t!]
\centering
\includegraphics[width=\linewidth,clip]{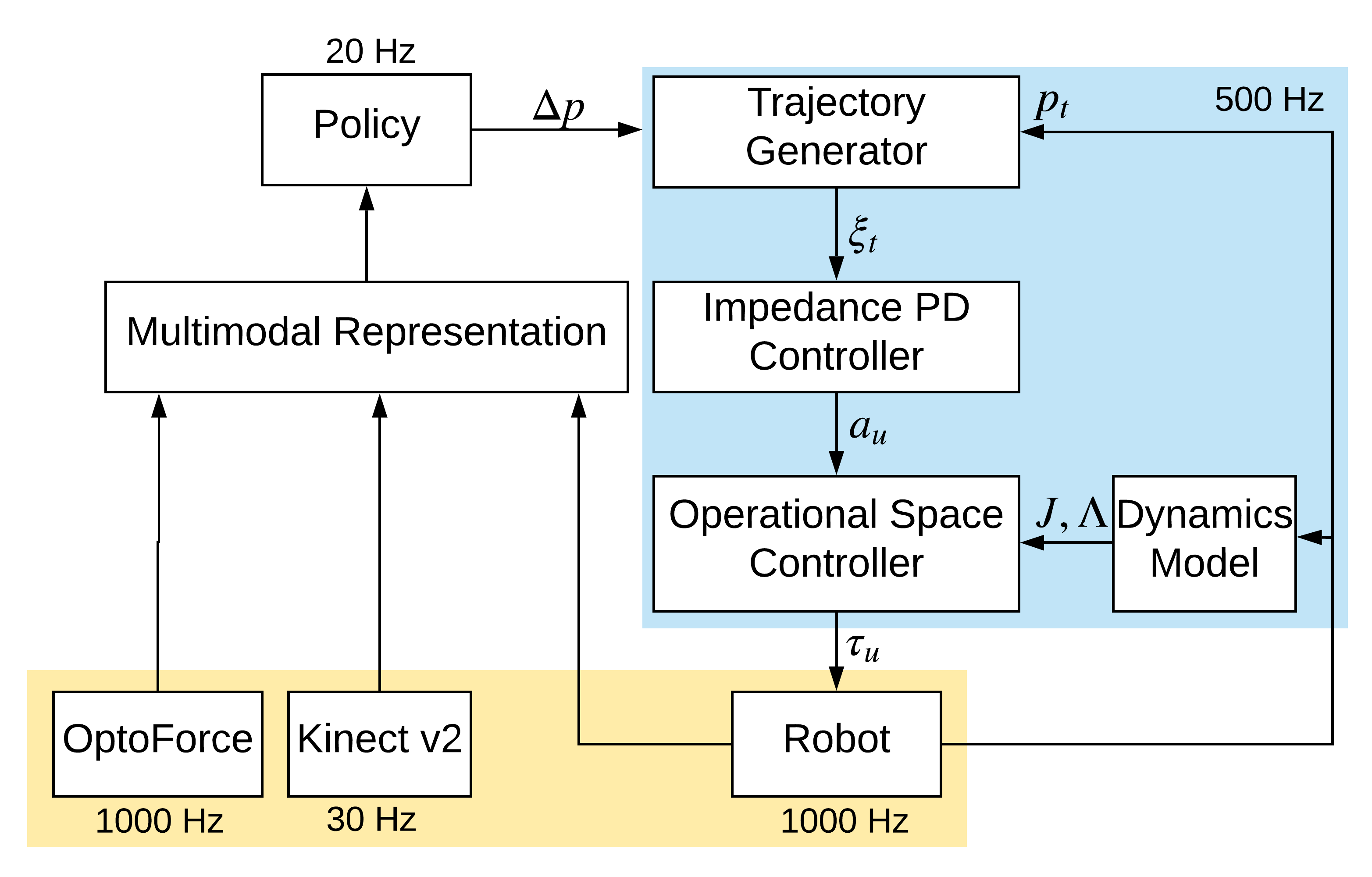}
\caption{Our controller takes end-effector position and z-axis orientation displacements from the policy at 20\unit{Hz} and outputs robot torque commands at 1000\unit{Hz}. The trajectory generator interpolates high-bandwidth robot trajectories from low-bandwidth policy actions. The impedance PD controller tracks the interpolated trajectory. The operational space controller uses the robot dynamics model to transform Cartesian-space accelerations into commanded joint torques. The resulting controller is compliant and reactive.}
\label{fig:controller}
\end{figure}

\section{Baseline Representation Models}

In addition to our variational self-supervised multimodal representation model (referred to as \texttt{Full Model}, we also propose two representation learning baselines for comparison. 

\subsection{Deterministic Model}
\label{sec:deterministic}
The \texttt{Deterministic model} is based after the model proposed in our previous work \cite{lee2018making}, which does not use a probabilistic graphical model framework. Instead we are using deterministic encoders to learn the representation and deterministic decoders to predict the same self-supervised objectives. Each modality encoder outputs a feature vector with $2\times d$ dimensions, where $d=128$. We concatenate the four feature vectors from each modality, and pass the resulting vector through a 2-layer MLP to produce the final $d$-dimensional multimodal representation. The decoder architectures remain the same as the \texttt{Full Model}, and we use the same self-supervised prediction objectives (and corresponding losses).

\subsection{Reconstruction Model}
\label{sec:vae}
The \texttt{Reconstruction model} does not use our proposed self-supervised learning objectives, but instead uses unsupervised learning to reconstruct the input modalities. In other words, it is a variational autoencoder ~\cite{kingma2013auto} using the same product of experts ~\cite{WuandGoodman} assumption and associated equations to combine the output parameters from the individual modality encoders. The model is trained to reconstruct the RGB image (vision), force reading (force) and end-effector pose and velocities (proprioception) inputted into the model. The loss function used to measure the error in the reconstruction for all three modalities is the mean squared error between the estimated values and the ground-truth values. 

The vision reconstruction decoder consists of 6 2D convolutional layers with upsampling, with the final hidden layer transformed with sigmoid activation. The proprioception decoder is composed of a 4-layer MLP. While the force modality's input to the network is a $32\times 6$ time series of force readings, the force decoder only estimates the force reading at the final timestep, instead of reconstructing the full time series array. Lastly, we use a 4-layer MLP to estimate the 6-dimensional force reading.

\section{Policy Learning and Controller Design}
\label{sec:policy-control}

Our final goal is to equip a robot with a policy for performing contact-rich manipulation tasks that leverages multimodal feedback. Though it is possible to engineer controllers for specific instances of these tasks~\cite{Whitney:1987,Song:2014}, this effort is difficult to scale to the large variability of real-world tasks. Therefore, it is desirable to enable a robot to learn control policies through trial-and-error, where the learning process is applicable to a broad range of tasks. In this work, we use a peg insertion task with different geometries as our evaluation task. 

Given its recent success in continuous control~\cite{lillicrap2015continuous,schulman2015trust}, deep reinforcement learning lends itself well to learning policies that map high-dimensional features to control commands. 

\noindent \textbf{Policy Learning.}
Modeling contact interactions and multi-contact planning still result in complex optimization problems~\cite{Posa:2013ez,ponton2016,tonneau18} that remain sensitive to inaccurate actuation and state estimation. We formulate contact-rich manipulation as a model-free reinforcement learning problem to investigate its performance when relying on multimodal feedback and when acting under uncertainty (such as uncertain geometry, clearance, and configuration for our peg insertion task). in 
By choosing model-free, we also eliminate the need for an accurate dynamics model, which is typically difficult to obtain in the presence of rich contacts. Specifically, we choose trust-region policy optimization (TRPO), which is a policy gradient method~\cite{schulman2015trust}. TRPO imposes a bound of KL-divergence for each policy update by solving a constrained optimization problem, which prevents the policy from moving too far away from the previous step. 
The policy network is a 2-layer MLP that takes as input the $d$-dimensional multimodal representation and produces 3D position displacement $\cartdelta$ and 1D orientation displacement $\rzdelta$ of the robot end-effector. To train the policy efficiently, we freeze the representation model parameters during policy learning, such that it reduces the number of learnable parameters to $1.5\%$ of the entire model and substantially improves the sample efficiency.


\noindent \textbf{Controller Design.} We define the 6-DoF pose of the end-effector $\mathbf{p}$ as consisting of end-effector position $\cartpos \in \mathbb{R}^3$ and end-effector orientation $\mathbf{R} \in \mathbf{SO}(3)$. Assuming the Euler Angle representation of rotation, we can define end-effector rotation around the fixed unit vectors of the global frame $(x,y,z)$ as $(\rx, \ry, \rz)$. In this work, we control the 3D position $\cartpos$ and the $z$-axis roll rotation $\rz$ of the end-effector (but do not control the $x$-axis yaw rotation $\rx$ and y-axis pitch rotation $\ry$). 

Our controller takes as input Cartesian end-effector position displacements $\cartdelta$ and roll angle displacements $\rzdelta$  from the policy at 20\unit{Hz}, and outputs direct torque commands $\taurobot$ to the robot at 500\unit{Hz}. The controller architecture can be split into three parts: trajectory generation, impedance control and operational space control (see Fig~\ref{fig:controller}). Our policy outputs end-effector control commands instead of joint-space commands, so it does not need to implicitly learn the non-linear and redundant mapping between 7-DoF joint space and 4-DoF end-effector space. We use direct torque control as it gives our robot compliance during contact, which makes the robot safer to itself, its environment, and any nearby human operator. In addition, compliance makes the peg insertion task easier to accomplish under position uncertainty, as the robot can slide on the surface of the box while pushing downwards \cite{Mrinal:2011,Righetti2014,Eppner:2015:EEC:2879361.2879370}.

\begin{figure*}[ht!]
    \centering
    \includegraphics[trim={0.5cm 6cm 0cm 2cm},clip, width=.99\linewidth]{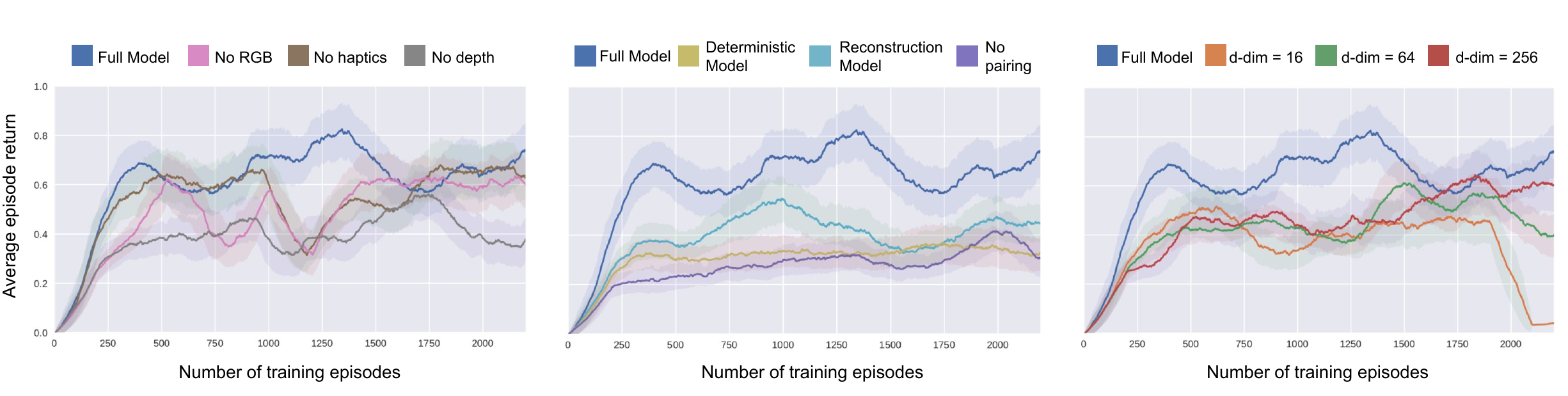}
    \caption{Reinforcement Learning Curves for representation trained on (A) different combinations of sensory modalities, (B) different loss functions, and (C) different representation latent space dimensions.}
    \label{fig:training_curves}
\end{figure*}

 The trajectory generator bridges low-bandwidth output of the policy (limited by the forward pass of our representation model), and the high-bandwidth torque control of the robot. Given $\cartdelta$ and $\rzdelta$ from the policy (and initial yaw angle $\rxinit$ and pitch angle $\ryinit$), we can construct a 6D end-effector displacement pose $\posedelta$. With a current 6D end-effector pose $\poset$, we calculate the desired end-effector pose  $\posedes$. The trajectory generator interpolates between $\poset$ and $\posedes$ to yield a trajectory $\xi_t = \{\poset, \cartveldt, \cartaccdt \}_{k=t}^{t+T}$ of end-effector pose, velocity and acceleration at 500\unit{Hz}. This forms the input to a PD impedance controller to compute a task space acceleration command:
$\cartaccu = \cartaccd -\kp (\cartpos-\cartposd) - \kv (\cartvel - \cartveld),$ where $\kp$ and $\kv$ are manually tuned gains.

By leveraging known kinematic and dynamics models of the robot, we can calculate joint torques from Cartesian space accelerations with the dynamically-consistent operational space formulation~\cite{Khatib1995a}. We compute the force at the end-effector with $\mathbf{f}= \Lambda \cartaccu$, where $\Lambda$ is the inertial matrix in the end-effector frame that decouples the end-effector motions. Finally, we map from $\mathbf{f}$ to joint torque commands with the end-effector Jacobian $\mathbf{J}$, which is a function of joint angle $\mathbf{q}$: $\taurobot = \mathbf{J}^T(\mathbf{q})\mathbf{f}$.

\section{Experiments: Design and Setup}
\label{sec:exp}

The primary goal of our experiments is to examine the effectiveness of the multimodal representations in contact-rich manipulation tasks. 
In particular, we design the experiments to answer the following five questions: 1) What is the value of using {\em all\/} instead of a subset of modalities? 2) What representation learning loss functions help policy learning? 3) How compact can the latent representation be for policy learning?  4) Is policy learning on the real robot \textit{practical} with a learned representation? 5) Does the learned representation \textit{generalize} over task variations and recover from perturbations?

\noindent \textbf{Task Setup.} 
We design a set of peg insertion tasks where task success requires joint reasoning over visual and haptic feedback. 
We use four different types of pegs and holes fabricated with a 3D printer: square peg, triangular peg, semicircular peg, and hexagonal peg, each with a nominal clearance of around 2mm as shown in \figref{peg_types}. 


\noindent \textbf{Robot Environment Setup.} 

In simulation, we use the Kuka LBR IIWA robot, a 7-DoF torque-controlled robot. In our previous work \cite{lee2018making}, we have used the same robot for real world experiments. Here, we use the Franka Panda robot (also with 7-DoF, torque-controlled) to emphasize that the results reported in \cite{lee2018making} are reproducible on different hardware. Four sensor modalities are available in both simulation and real hardware, including proprioception, an RGB-D camera, and a force-torque sensor. The proprioceptive input is the end-effector pose as well as linear and angular velocity. They are computed using forward kinematics. RGB images and depth maps are recorded from a fixed camera pointed at the robot. Input images to our model are down-sampled to $128\times 128$. On the real robot, we use the Kinect v2 camera. In simulation, we use CHAI3D~\cite{Conti03} for rendering images and robot meshes for rendering depths \cite{GarciaCifuentes.RAL}. The force sensor provides 6-axis feedback on the forces and moments along the x, y, z axes. On the real robot, we mount an OptoForce sensor between the last joint and the peg. In simulation, the contact between the peg and the box is modeled with SAI 2.0~\cite{conti2016framework}, a real-time physics simulator for rigid articulated bodies with high fidelity contact resolution. 

\noindent \textbf{Reward Design.}
We use the following staged reward function to guide the reinforcement learning algorithm through the different sub-tasks, simplifying the challenge of exploration and improving learning efficiency:
%
\[
r(\mathbf{s}) =
  \begin{mcases}[l@{\ }]
    c_r(1-(\tanh{\lambda\|\mathbf{s}\||})(1-s_{\psi}) &\text{(r)}\\
    
    1 + c_a(1 - \frac{\|\mathbf{s}\|_2}{\|\mathbf{\epsilon}_{1}\|_2})(1 - \frac{s_{\psi}}{\epsilon_{\psi}}) \qquad\; \text{if} \;  \mathbf{s} \leq \mathbf{\epsilon}_{1}\And s_{\psi} \leq \epsilon_\psi & \text{(a)} \\
    
    2 + c_i(h_d - \|s_z\|) \qquad\qquad\qquad\; \text{if} \;  s_z < 0 &\text{(i)} \\
    
    5 \qquad\qquad\qquad\qquad\qquad\qquad\; \text{if} \;  h_d - |s_z| \leq \epsilon_2&\text{(c)}
  \end{mcases}
\]

where $\mathbf{s}=(s_x, s_y, s_z)$ denotes the peg's current relative position to the peg hole and $s_{\psi}$ is the current relative orientation along the z axis of the peg in relation to the peg hole, $\lambda$ is a constant factor to scale the input to the $\tanh$ function. The target peg position is $(0,0,-h_d)$ with $h_d$ as the height of the hole, and $c_r$ and $c_a$ are constant scale factors.

\noindent \textbf{Evaluation Metrics.} We report the quantitative performance of the policies using the sum of rewards achieved in an episode, normalized by the highest attainable reward. We also provide statistics on the stages of the peg insertion task that each policy can achieve, and report the percentage of evaluation episodes in the following four categories:
\begin{enumerate}
    \item \emph{completed insertion}: peg reaches bottom of the hole;
    \item \emph{inserted into hole}: peg goes into the hole but has not reached the bottom;
    \item \emph{touched the box}: peg only makes contact with the box; 
    \item \emph{failed}: peg fails to reach the box.
\end{enumerate}
\begin{figure*}[t!]
    \vspace{-10pt}
    \centering
    \includegraphics[trim={1cm 1cm 0cm 2cm},clip,width=\linewidth]{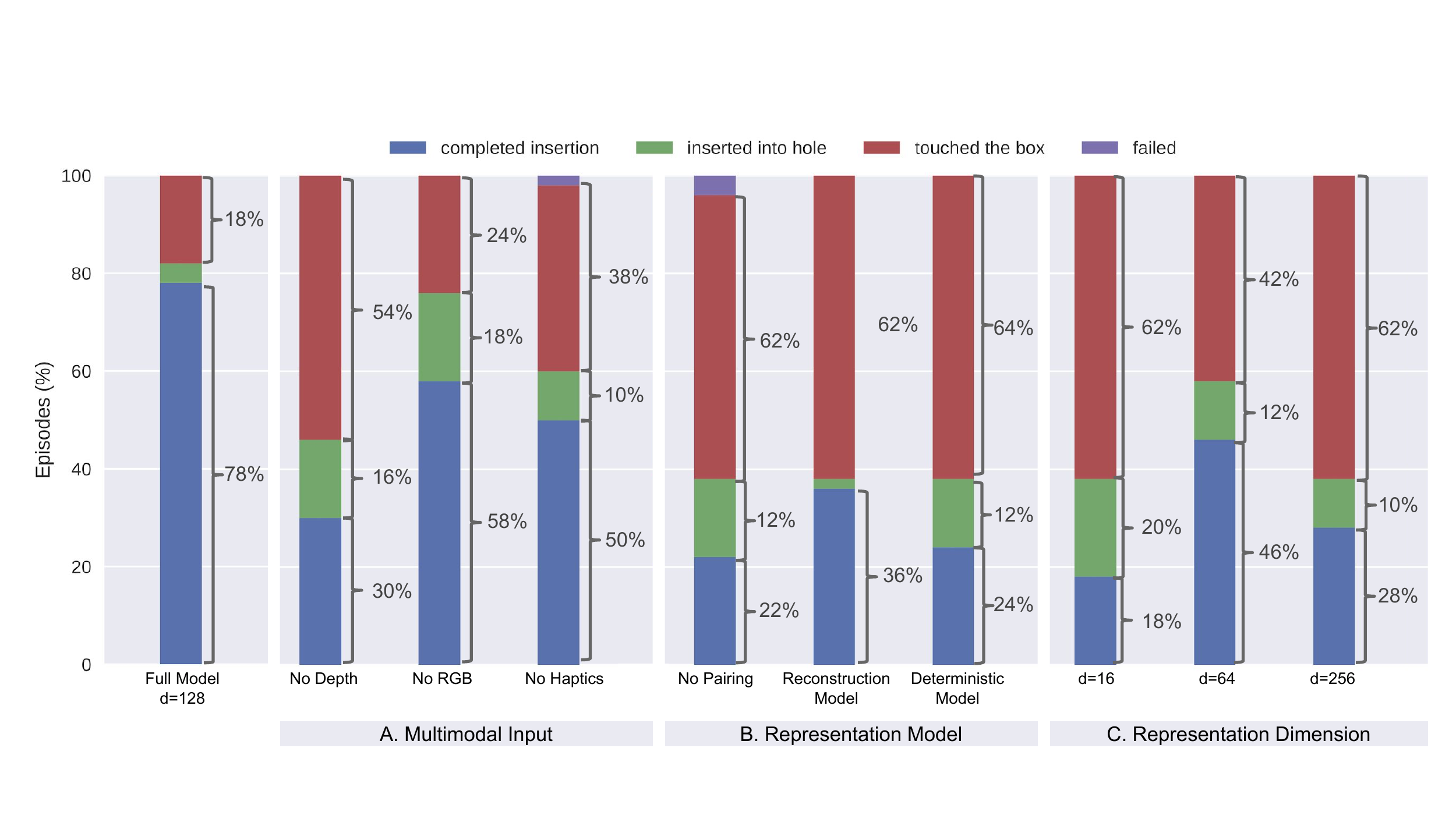}    
     \caption{Simulated Peg Insertion: Ablative study of representations trained on (A) different combinations of sensory modalities, (B) different representation models, and (C) different representation latent space dimensionality. In (A), we compare our Full Model, trained on vision, haptic and proprioceptive data, with baselines that are either trained without RGB images, depth, or haptics. The graph shows partial task completion rates with different feedback modalities, and we note that depth, RGB images, and haptic modalities play an integral role for contact-rich tasks. In (B), we note a performance drop of alternative representation models, suggesting the pairing training objective, action-conditional predictions, and variational inference are all important training objectives in learning a representation model. In (C), by varying the dimension of the representation model, we see that policy performance drops when the latent space dimension is too small to capture important state information. When the latent space dimension is too large, policy learning also suffers.}
    \label{fig:ablation_stats}
\end{figure*}

\noindent \textbf{Implementation Details.} 
To train each representation model, we collect a multimodal dataset of 150k states on a real robot and 600k in simulation. We generate the self-supervised annotations offline. We roll out a random policy as well as a heuristic policy while collecting the data, which encourages the peg to make contact with the box. As the policy runs at 20 Hz, 100k data takes around 90 minutes to collect. The representation models are trained for 20 epochs on a Quadro P5000 GPU before starting policy learning. Details of how we train our representation and policy can be found in Appendix \ref{appendix:rep} and Appendix \ref{appendix:rl} respectively.

\section{Experiments: Results}
We first conduct an ablative study in simulation to investigate the contributions of individual sensory modalities, representation learning techniques, and latent space dimensionality to learning the multimodal representation and manipulation policy. We then apply our full multimodal model to a real robot, and train policies using reinforcement learning for the peg insertion tasks from the learned representations with high sample efficiency.




\subsection{Simulation Experiments}

\subsubsection{Multimodal Input Experiments}
 Four modalities are encoded and fused by our representation model: RGB images, depth images, force readings, and proprioception (see Fig.~\ref{fig:network_architecture}). To investigate the importance of each modality for contact-rich manipulation tasks, we perform an ablative study in simulation where we learn multimodal representations with different combinations of modalities. These learned representations are subsequently fed to the policy networks to train on a task of inserting a square peg. While we use a probabilistic representation during representation learning, we take the mean of the learned representation during policy learning. We randomize the configuration of the box position at the beginning of each episode to enhance the robustness and generalization of the model.

We illustrate the training curves of the TRPO-trained agents in \figref{training_curves}. We train all policies with 2.0k episodes, each lasting 500 steps. We updated the policy networks every four episodes. To evaluate the policies, we chose the best performing checkpoint that was logged during training for each policy based on the training results and performed 50 rollouts on each policy. The results of the evaluation can be seen in \figref{ablation_stats}.

Our \texttt{Full Model} corresponds to the multimodal representation model introduced in \secref{representation-learning}, which takes all four modalities as input. We compare it with three baselines: \texttt{No RGB} masks out the visual input to the network, \texttt{No Haptics} masks out the haptic input, and \texttt{No Depth} masks out the depth input.
From \figref{training_curves} and \figref{ablation_stats} we observe that the absence of the RGB images, depth, or force modality negatively affects task completion, with \texttt{No Depth} performing the worst. Among these three baselines, we see that the \texttt{No RGB} baseline achieved the highest rewards, suggesting that a combination of visual data from depth and haptics data from the force sensor gives sufficient information for the peg insertion tasks. None of the three baselines have reached the same level of performance as the final model, which uses all the modalities,



\subsubsection{Representation Learning Model}
\label{sec:replearningmodel}
Our multimodal representation \texttt{Full Model}, as described in Sec.~\ref{sec:representation-learning}, uses variational encoders to predict action-conditional optical flow, contacts, end-effector pose, and time-aligned sensory pairing. We further investigate the efficacy of our model by comparing it to three baselines: \texttt{Deterministic Model} using deterministic encoders (and trained without ELBO loss) as described in Sec. \ref{sec:deterministic}, \texttt{No Pairing} that is trained without the sensory time-alignment prediction, and \texttt{Reconstruction Model} as described in Sec. \ref{sec:vae}. Similar to the modality input ablation study, these learned representation models are subsequently fed to policies trained to insert a square peg using TRPO. 
As stated earlier, our \texttt{Full Model} completes insertion 78\% of the time. In \cite{lee2018making}, the \texttt{Deterministic Model} completed insertion at 76\% success rate for a 3-DoF peg insertion task. With the additional orientation control, the \texttt{Deterministic Model} task success rate drops to 24\%. This drop in performance demonstrates the challenge of learning an insertion task with both rotation and translation actions, as well as the efficacy of the probabilistic encoder. Recent work that studies deterministic and variational inference approaches have shown that variational inference regularizes learning by enforcing a smooth latent space structure \cite{ghosh2019variational}. In our work, we see signs of the \texttt{Deterministic Model} overfitting. Compared to the training losses, the test losses for contact prediction and pairing prediction increase by a factor of 312.11 and 2514 respectively (see Table \ref{tab:ratio}).

\begin{table}[t!]
\footnotesize
\caption{Ratio of Representation Test over Training Loss for Full Model (FM), Deterministic Model (Det), No Pairing (NP), d=256 (256), d=64 (64), and d=16 (16)}
\label{tab:ratio}
\begin{tabular}{lllllll}
\hline
Loss              & FM & Det & NP & 256 & 64 & 16 \\ \cline{1-7}
Optical Flow      & 1.63&	1.79&	1.26&	1.27&	1.27&	1.81     \\
End-Effector Pose & 0.75       & 0.91          & 0.56       & 0.32      & 0.31     & 1.96     \\
Contact           & 13.84      & 312.11        & 0.68       & 0.87      & 0.87     & 2.51     \\
Pairing           & 4.99       & 2,514.26      & N/A        & 26.00     & 11.31    & 32.29    \\ \hline
\end{tabular}
\end{table}

We observe that our self-supervised training objectives are important for achieving the \texttt{Full Model} performance, especially the time-alignment pairing loss. We see that \texttt{No Pairing} affects the policy learning the most, with insertion rates dropping to 22\%. According to~\cite{owens2018audio}, deciding whether sensory streams are time-aligned requires the detection of co-occuring patterns across modalities. These patterns provide evidence for a common underlying event, e.g. making or breaking contact. The importance of the pairing loss for task performance suggests that learning these patterns that co-occur between modalities provides a strong learning signal. The policy that uses \texttt{Reconstruction Model} representation learns at a faster rate and performs more insertions than the \texttt{Deterministic Model} and \texttt{No Pairing}. However, the full insertion rate of 36\% is still less than half of the insertion rate of the \texttt{Full Model}. For contact-rich tasks, our action-conditional, self-supervised objectives are easier to learn than the full reconstruction objectives, and also are more suitable for policy learning.  

\subsubsection{Representation Dimensions}
\label{sec:size}
We evaluate how compact the representation needs to be for contact-rich manipulation task by changing the dimensionality of the multimodal representation. We hypothesize that while a more compact representation can make reinforcement learning more tractable, it also captures less information about the state. We test several dimensions: \texttt{d=16}, \texttt{d=64}, \texttt{d=128} (our \texttt{Full Model}), and \texttt{d=256}. We see that the \texttt{d=16} model can only fully insert 18\% of the time. It also has the lowest training accuracy for contact prediction (95.3\%) and highest end-effector pose prediction loss (1.56E-02) compared to the other representation dimension sizes (see Table \ref{tab:ratio} and Appendix \ref{appendix:rep}, Table \ref{tab:replearning}). This suggests that the model captures too little information about the state for the task. While \texttt{Full Model} performs the best with 78\% full insertion, the performance of the policy drops by more than a third when the representation size increases to \texttt{d=256}. As seen in Table \ref{tab:ratio}, \texttt{d=256} has a lower ratio between test loss and training loss for all the predictions except for pairing loss, with comparable absolute losses in each category (as seen in the Appendix \ref{appendix:rep}, Table \ref{tab:replearning}), this suggests that the model learned the predictions well with little overfitting. The drop in performance in policy learning for \texttt{d=256} might be due to the increase of the state space for the policy. 

\begin{figure}
    \centering
    \includegraphics[width=1.0\linewidth]{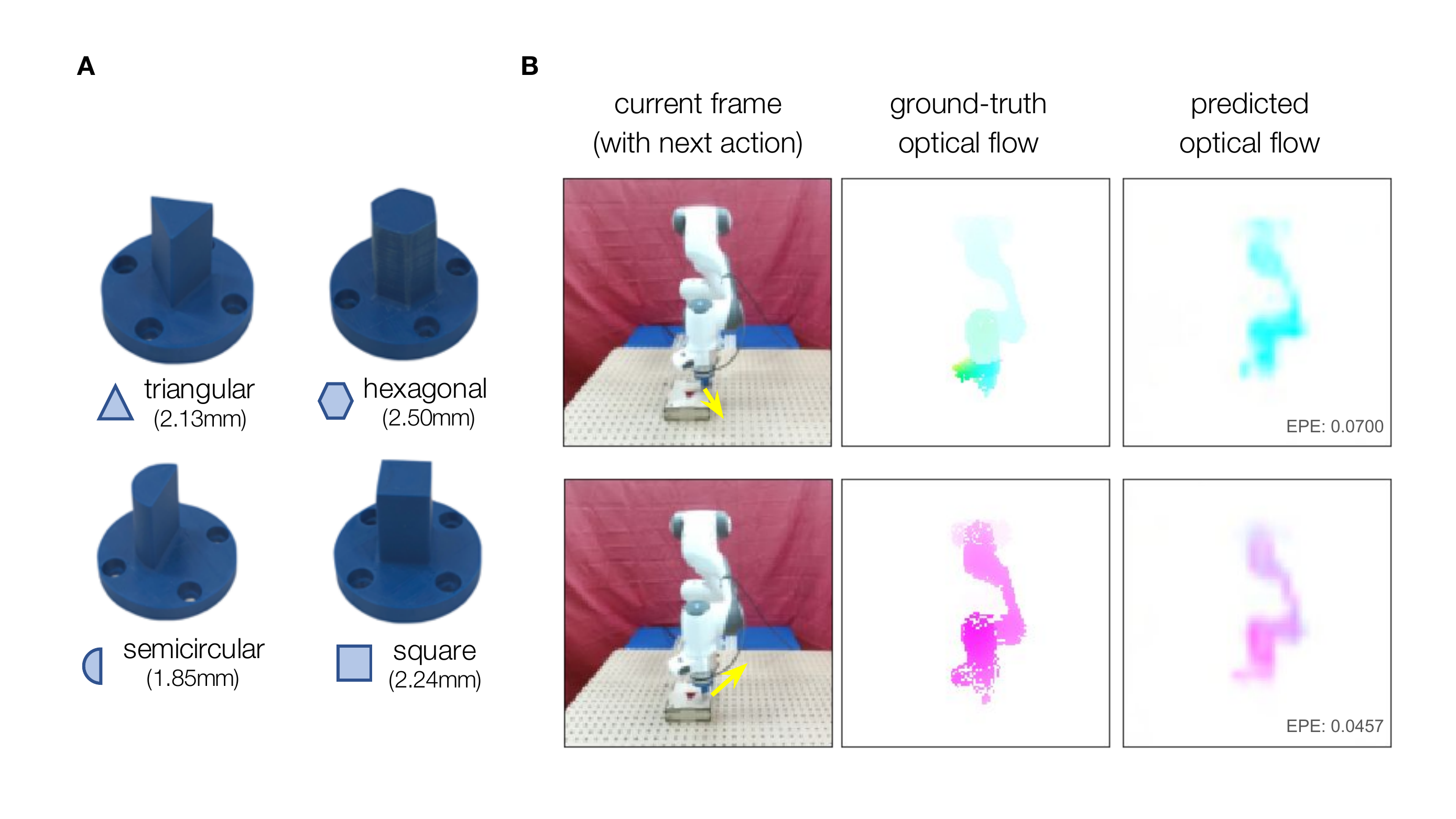}
    \caption{(A) 3D printed pegs used in the real robot experiments and their box clearances. (B) Qualitative predictions: We visualize examples of optical flow predictions from our representation model (using color scheme in~\cite{flownet1}). The model predicts different flow maps on the same image conditioned on different next actions indicated by projected arrows.}
    \label{fig:peg_types}
\end{figure}

\subsection{Real Robot Experiments}
In our previous work \cite{lee2018making}, we evaluated the \texttt{Deterministic Model} with a 3D action space representing Cartesian position displacements. On the physical robot platform we evaluated the policies with round, triangular, and semicircular pegs. In this work, we evaluate our \texttt{Full Model} on the real hardware with triangular and semicircular pegs only, since the circular peg does not require orientation control for insertion. In contrast to simulation, the difficulty of sensor synchronization, variable delays from sensing to control, and complex real-world dynamics introduce additional challenges. We make the task tractable on a real robot by training a shallow neural network controller after freezing the multimodal representation model when it is able to generate action-conditional flows with low endpoint errors (see \figref{peg_types}). 

\begin{figure}
    \centering
    \includegraphics[trim={2cm 0cm 1.5cm 0.5cm},clip,width=\columnwidth]{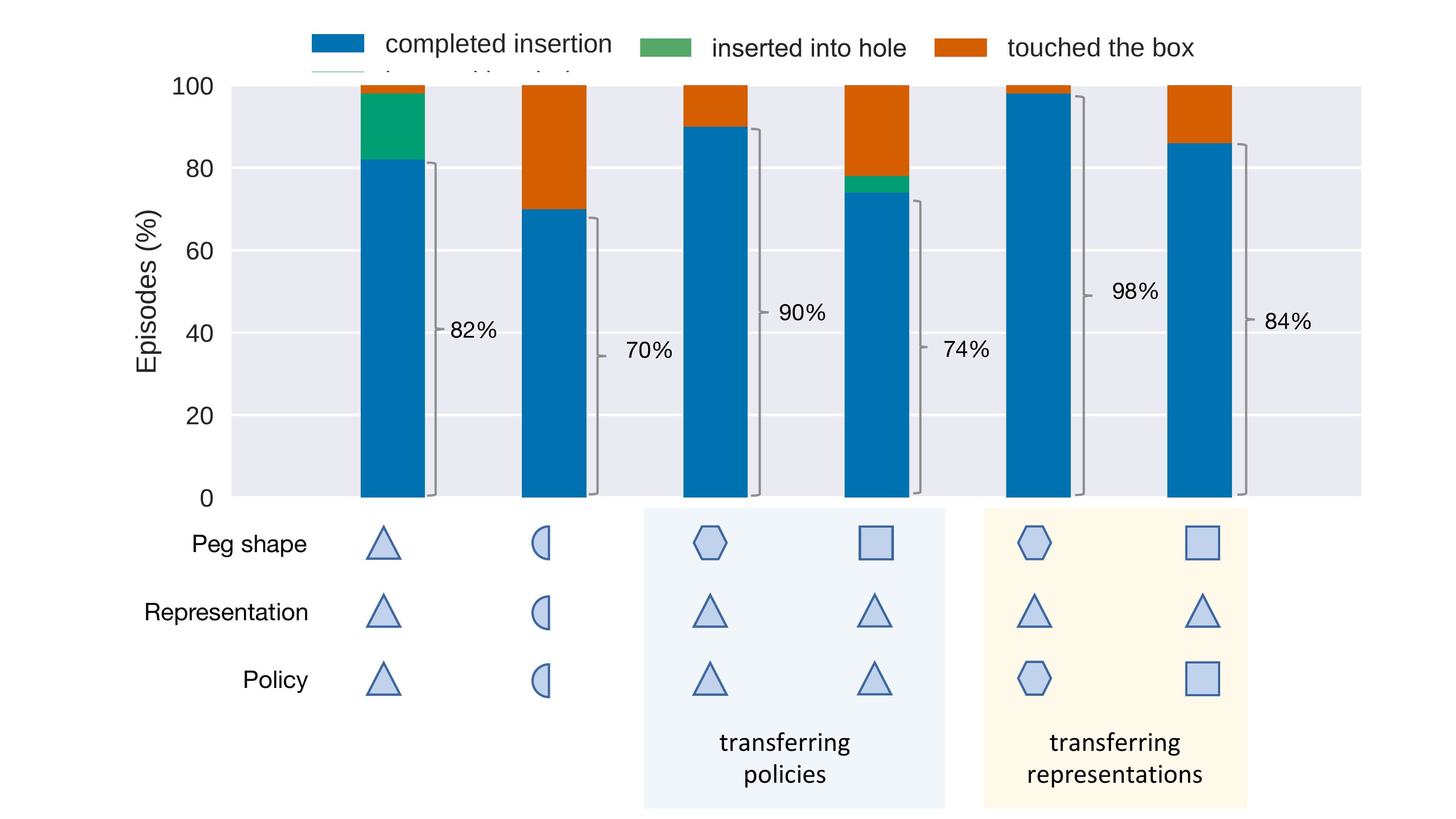}
    \caption{Real Robot Peg Insertion: We evaluate our \texttt{Full Model}
    on the real hardware with different peg shapes, indicated on the x-axis. The learned policies achieve the tasks with a high success rate. We also study transferring the policies and representations from trained pegs to novel peg shapes (last four bars). The robot effectively re-uses previously trained models to solve new tasks.}
    \label{fig:real_robot_exp}
\end{figure}

We train the policy networks for 450 episodes, each lasting 1000 steps, roughly 7 hours of wall-clock time. We evaluate each policy for 50 episodes in \figref{real_robot_exp}. The first two bars correspond to the set of experiments where we train a specific representation model and policy for each type of peg. The robot achieves a level of success similar to that in simulation. A common strategy that the robot learns is to reach the box, search for the hole by sliding over the surface, align the peg with the hole, and finally perform insertion. More qualitative behaviors can be found in the supplementary video.

We further examine the potential of transferring the learned policies and representations to two novel shapes previously unseen in representation and/or policy learning: the hexagonal peg and the square peg. For policy transfer, we take the representation model and the policy trained for the triangular peg, and test it with the new pegs. From the 3rd and 4th bars in \figref{real_robot_exp}, we see that the policy achieves over 70\% success rate on both pegs without any further policy training on them. A better transfer performance can be achieved by taking the representation model trained on the triangular peg, and training a new policy for the new pegs. As shown in the last two bars in \figref{real_robot_exp}, the resulting performance increases by 8\% for the hexagonal peg and by 10\% for the square peg. Our transfer learning results indicate that the multimodal representations from visual and haptic feedback generalize well across geometric variations of our contact-rich manipulation tasks.

Finally, we study the robustness of our policy in the presence of sensory noise and external perturbations to the arm by periodically occluding the camera and pushing the robot arm during trajectory roll-out. The policy is able to recover from both the occlusion and perturbations. Qualitative results can be found in our supplementary video on our website: \url{https://sites.google.com/view/visionandtouch}.


\section{Discussion and Conclusion}


We examined the value of learning a joint representation of time-aligned multisensory data for contact-rich manipulation tasks. To enable efficient real robot training, we proposed a novel model to encode heterogeneous sensory inputs into a compact multimodal latent representation. Once trained, the representation remained fixed when being used as input to a shallow neural network policy for reinforcement learning. We trained the representation model with self-supervision, eliminating the need for manual annotation. Our experiments with tight clearance peg insertion tasks indicated that they require the multimodal feedback from both vision (RGB and depth) and touch. We also showed that models trained with our proposed self-supervised action-conditional prediction and time-alignment pairing prediction objective surpass models trained on reconstruction objectives. Our ablation studies show that the pairing prediction objective during representation learning is especially important for policy performance, as the prediction allows our representation to learn the relationship between the sensor modalities.  By varying the dimension of the latent space representation, we observed that a larger latent space can better learn the self-supervised objectives. When the latent space is too large, it can adversely affect the policy learning. In other words, the size of the latent space is a trade-off between capturing enough information of the state and keeping the policy state space compact. It would be beneficial to study more principled methods of making this trade-off. 

On the real robot, we demonstrated that the multimodal representations transfer well to new task instances of peg insertion. For future work, we plan to extend our method to other contact-rich tasks, which require a full 6-DoF controller of position and orientation. 
We would also like to explore the value of incorporating richer modalities, such as sound, temperature, and proximity sensors, into our representation learning pipeline, as well as new sources of self-supervision.

\section{Acknowledgements}

This work has been partially supported by JD.com American Technologies Corporation (“JD”) under the SAIL-JD AI Research Initiative and by the Toyota Research Institute ("TRI"). This article solely reflects the opinions and conclusions of its authors and not of JD, any entity associated with JD.com, TRI, or any entity associated with Toyota.


\printbibliography
\newpage
\begin{IEEEbiography}[{\includegraphics[width=1in,height=1.25in,clip,keepaspectratio]{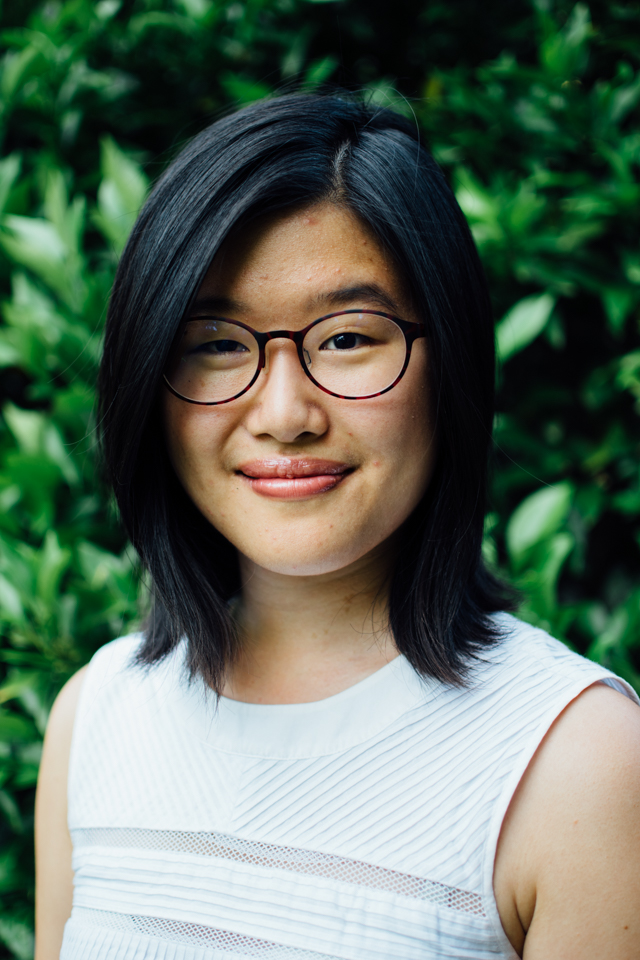}}]{Michelle A. Lee}
is a Ph.D. candidate in the Department of Mechanical Engineering at Stanford University, advised by Prof. Jeannette Bohg. She works on perception, controls, and machine learning for robot manipulation, and researches how representations of state and action spaces can enable algorithmic generalization, robustness, and efficiency. She received her Master's degree in Mechanical Engineering and Bachelor's degree in Chemical Engineering from Stanford University. 
\end{IEEEbiography}

\begin{IEEEbiography}[{\includegraphics[width=1in,height=1.25in,clip,keepaspectratio]{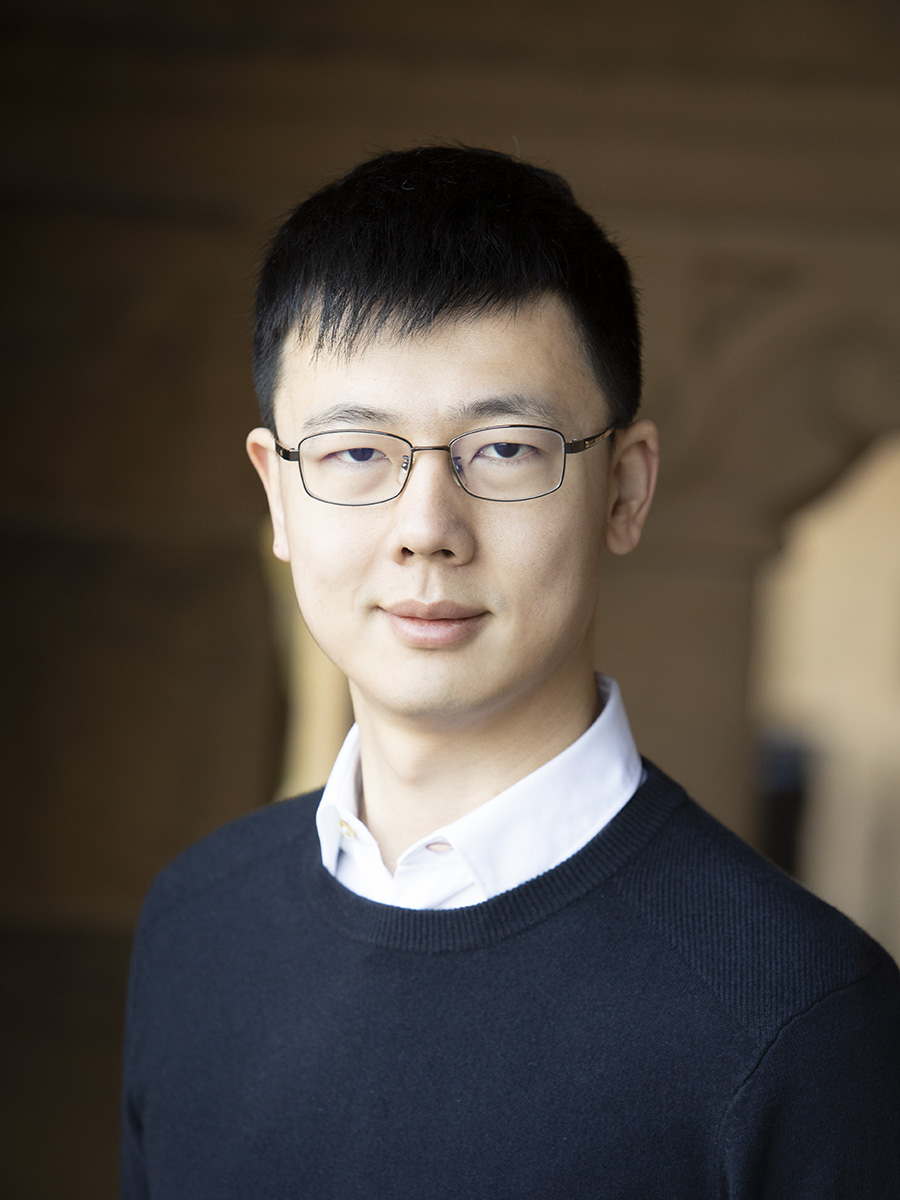}}]{Yuke Zhu}
is a final year Ph.D. candidate in the Department of Computer Science at Stanford University, advised by Prof. Fei-Fei Li and Prof. Silvio Savarese. His research interests lie at the intersection of robotics, computer vision, and machine learning. His work builds machine learning and perception algorithms for general-purpose robots. He received a Master's degree from Stanford University and dual Bachelor's degrees from Zhejiang University and Simon Fraser University. He also collaborated with research labs including Snap Research, Allen Institute for Artificial Intelligence, and DeepMind.
\end{IEEEbiography}

\begin{IEEEbiography}[{\includegraphics[width=1in]{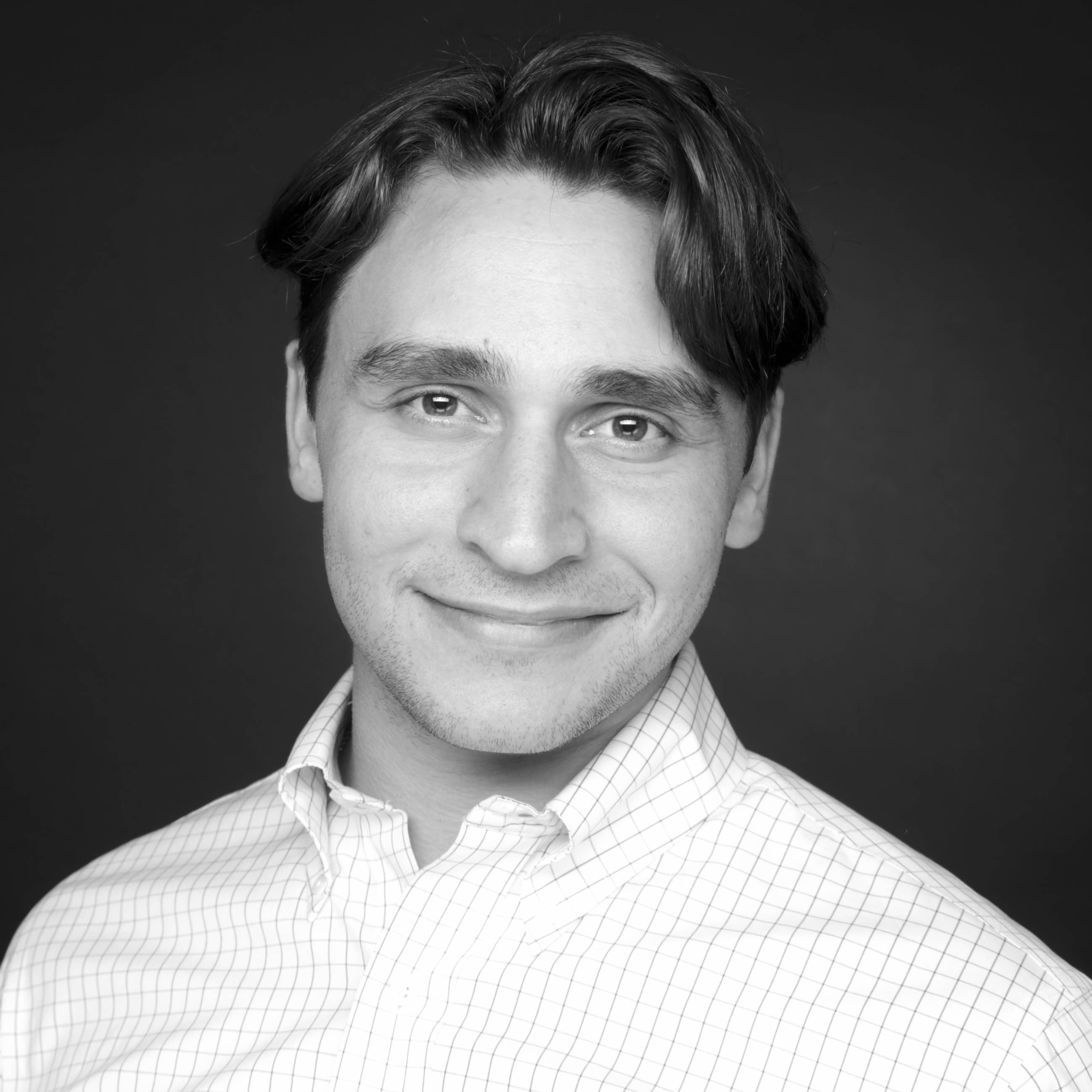}}]{Peter Zachares}
is a Master's student in the Department of Mechanical Engineering at Stanford University, advised by Prof. Jeannette Bohg. His work focuses on planning, controls, and machine learning for robot manipulation. He received his Bachelor's degree from the University of Pennsylvania. 
\end{IEEEbiography}

\begin{IEEEbiography}[{\includegraphics[width=1in,keepaspectratio]{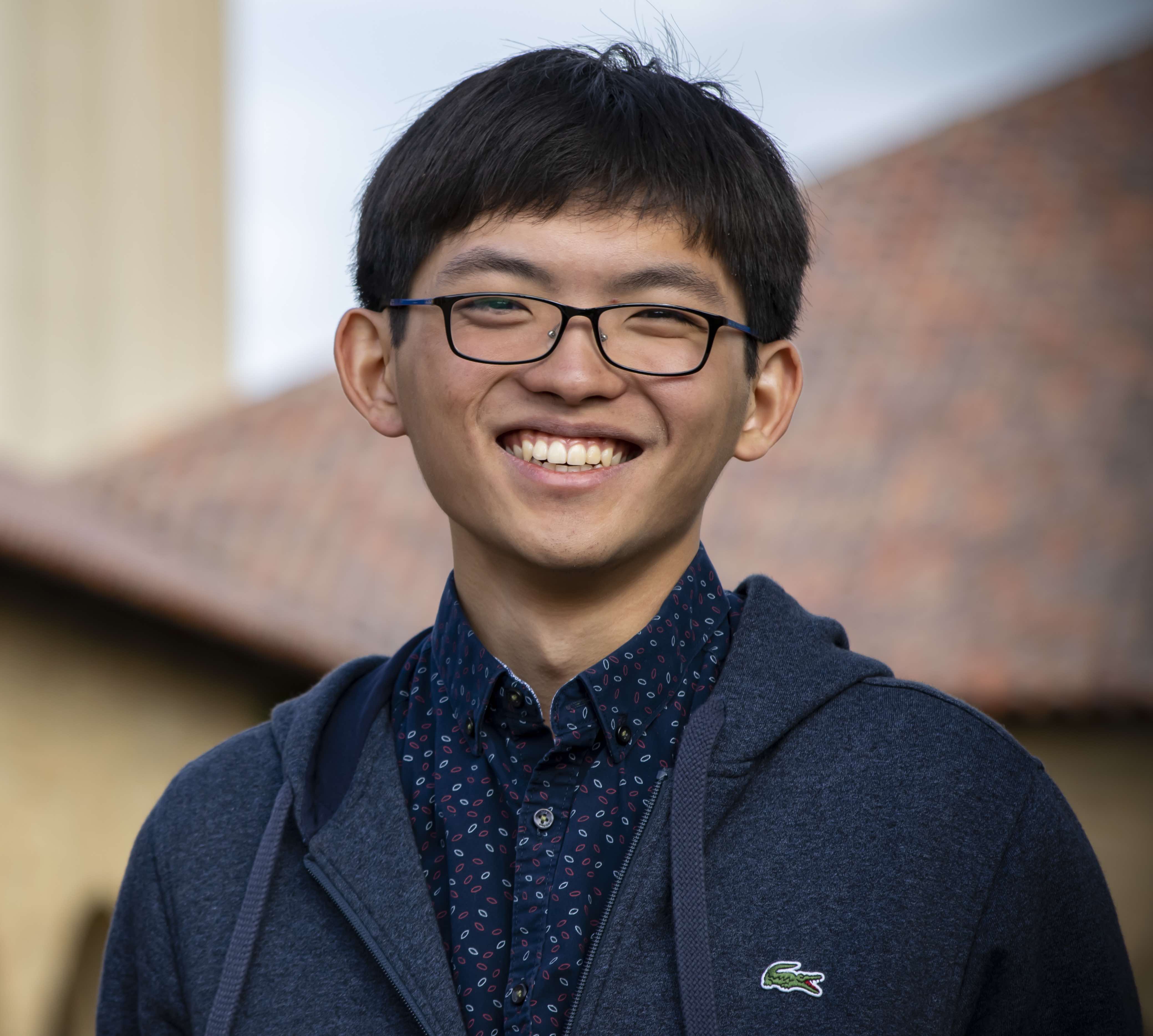}}]{Matthew Tan}
is an undergraduate student in the Department of Computer Science at Stanford University. He does research with Prof. Jeannette Bohg and is interested in machine learning, robot control and manipulation. 
\end{IEEEbiography}

\begin{IEEEbiography}[{\includegraphics[width=1in,keepaspectratio]{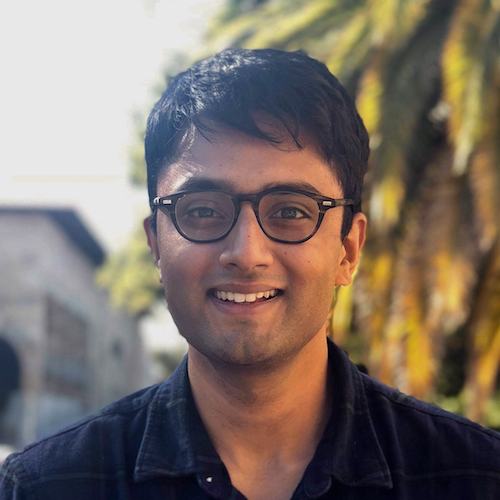}}]{Krishnan Srinivasan}
is a first-year Ph.D. student in the Department of Computer Science at Stanford University, advised by Prof. Jeannette Bohg. His current research focuses on robot learning, control, and in-hand manipulation, and he is interested in hierarchical reinforcement learning and meta-learning. He received his Bachelor's degree in Computer Science and Math at Yale University.
\end{IEEEbiography}


\begin{IEEEbiography}[{\includegraphics[width=1in,keepaspectratio]{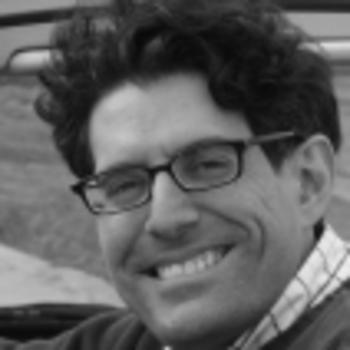}}]{Silvio Savarese}
 is an Associate Professor of Computer Science at Stanford University and the inaugural Mindtree Faculty Scholar. He earned his Ph.D. in Electrical Engineering from the California Institute of Technology in 2005 and was a Beckman Institute Fellow at the University of Illinois at Urbana-Champaign from 2005–2008. He joined Stanford in 2013 after being Assistant and then Associate Professor of Electrical and Computer Engineering at the University of Michigan, Ann Arbor, from 2008 to 2013. His research interests include computer vision, robotic perception and machine learning. He is recipient of several awards including a Best Student Paper Award at CVPR 2016, the James R. Croes Medal in 2013, a TRW Automotive Endowed Research Award in 2012, an NSF Career Award in 2011 and Google Research Award in 2010. In 2002 he was awarded the Walker von Brimer Award for outstanding research initiative.
\end{IEEEbiography}

\begin{IEEEbiography}[{\includegraphics[width=1in,keepaspectratio]{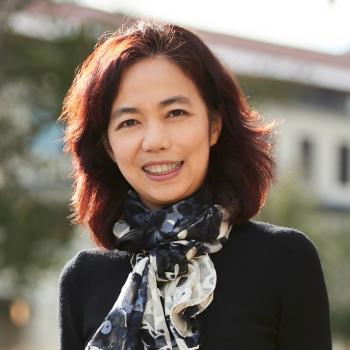}}]{Fei-Fei Li} is a Professor in the Computer Science Department at Stanford University, and Co-Director of Stanford’s Human-Centered AI Institute. She served as the Director of Stanford’s AI Lab from 2013 to 2018. And during her sabbatical from Stanford from January 2017 to September 2018, she was Vice President at Google and served as Chief Scientist of AI/ML at Google Cloud. Dr. Fei-Fei Li obtained her B.A. degree in physics from Princeton in 1999 with High Honors, and her PhD degree in electrical engineering from California Institute of Technology (Caltech) in 2005. She joined Stanford in 2009 as an assistant professor. Prior to that, she was on faculty at Princeton University (2007-2009) and University of Illinois Urbana-Champaign (2005-2006).\end{IEEEbiography}

\begin{IEEEbiography}[{\includegraphics[width=1in,height=1.25in,clip,keepaspectratio]{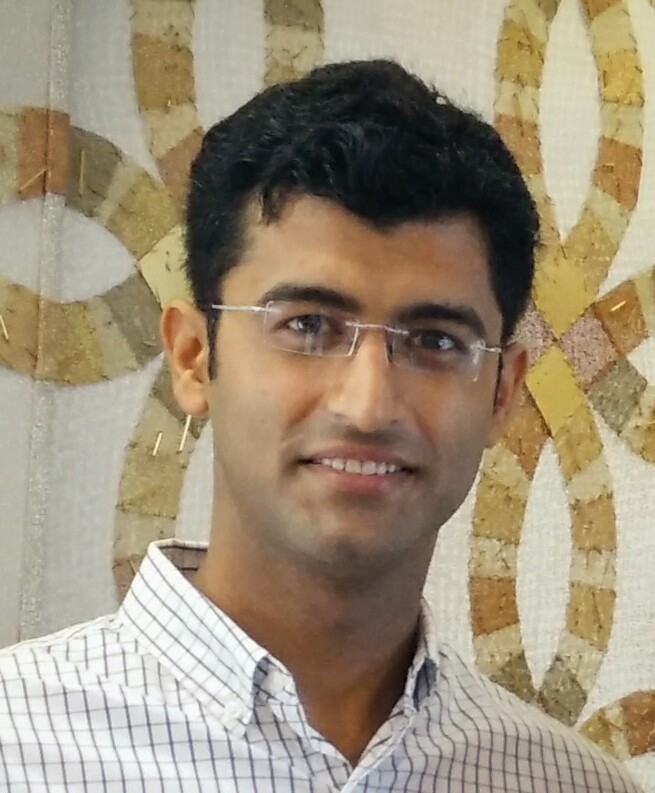}}]{Animesh Garg}
is an Assistant Professor of Computer Science at the University of Toronto. His research focus is to understand representations and algorithms to enable the efficiency and generality of learning for interaction in autonomous agents. He earned a Ph.D. from the University of California, Berkeley and was a postdoc at the Stanford AI Lab. He recevied a Master's degree from the Georgia Institute of Technology and a Bachelor's degree from the University of Delhi.
\end{IEEEbiography}

\begin{IEEEbiography}[{\includegraphics[width=1in,height=1.25in,clip,keepaspectratio]{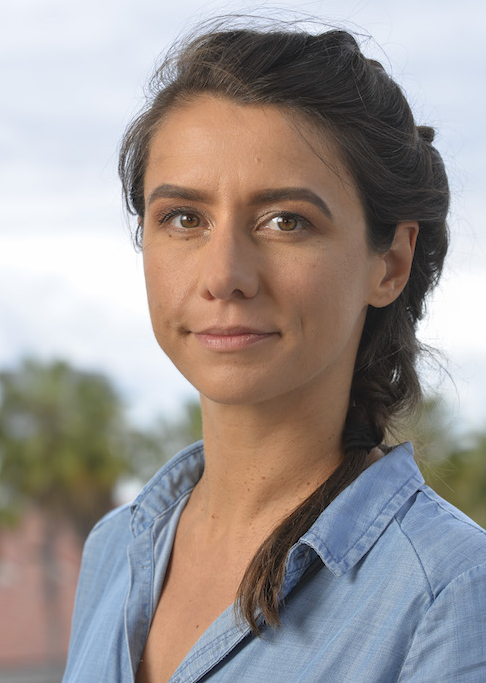}}]{Jeannette Bohg}
is an Assistant Professor of Computer Science at Stanford University. She was a group leader at the Autonomous Motion Department (AMD) of the MPI for Intelligent Systems until September 2017. Before joining AMD in January 2012, Jeannette Bohg was a PhD student at the Computer Vision and Active Perception lab (CVAP) at KTH in Stockholm. Her thesis on Multi-modal scene understanding for Robotic Grasping was performed under the supervision of Prof. Danica Kragic. She studied at Chalmers in Gothenburg and at the Technical University in Dresden where she received her Master in Art and Technology and her Diploma in Computer Science, respectively. Her research focuses on perception for autonomous robotic manipulation and grasping. She is specifically interesting in developing methods that are goal-directed, real-time and multi-modal such that they can provide meaningful feedback for execution and learning.
\end{IEEEbiography}

\clearpage

\newpage
\onecolumn

\begin{appendices}
\section{Representation Training Details}
\label{appendix:rep}

\FloatBarrier
\begin{table}[h!]
\caption{Hyperparameters for Representation Learning using Adam}
\label{tab:rephyper}
\begin{center}
\begin{tabular}{ll}
\hline

                         & All Models \\ \hline
Batchsize                & 64         \\
Learning Rate ($\alpha$) & 1.00E-04   \\
Adam Beta1 ($\beta_1$)   & 0.5        \\
Adam Beta2 ($\beta_2$)   & 0.999      \\ \hline
\end{tabular}
\end{center}
\end{table}
\FloatBarrier

For representation learning, we used Adam \cite{kingma2014adam} to run stochastic gradient descent over the prediction objectives as described in Sec. \ref{sec:loss}. These were the hyperparameters used during representation learning for both simulation and real robot data. 

\begin{table}[h!]
\caption{Representation Learning Prediction Losses}
\begin{center}

\label{tab:replearning}
\begin{tabular}{llllllll}
                \hline                   
& Dataset & Full Model & Deterministic & No Pairing & d-dim 256 & d-dim 64 & d-dim 16 \\ \hline
\multirow{2}{*}{Optical Flow Loss}      & Train   & 0.018      & 0.016         & 0.020      & 0.019     & 0.019    & 0.017    \\
                                        & Test    & 0.029      & 0.028         & 0.025      & 0.024     & 0.024    & 0.031    \\
\multirow{2}{*}{End-Effector Pose Loss} & Train   & 8.10E-03   & 1.76E-04      & 1.31E-02   & 1.66E-02  & 1.66E-02 & 1.56E-02 \\
                                        & Test    & 6.07E-03   & 1.60E-04      & 7.25E-03   & 5.34E-03  & 5.12E-03 & 3.05E-02 \\
\multirow{2}{*}{Contact Loss}           & Train   & 0.033      & 0.002         & 0.095      & 0.099     & 0.099    & 0.032    \\
                                        & Test    & 0.459      & 0.657         & 0.065      & 0.086     & 0.086    & 0.080    \\
\multirow{2}{*}{Contact Accuracy}       & Train   & 98.4\%     & 100.0\%       & 98.4\%     & 98.4\%    & 96.9\%   & 95.3\%   \\
                                        & Test    & 98.9\%     & 99.9\%        & 99.2\%     & 99.5\%    & 97.8\%   & 98.6\%   \\
\multirow{2}{*}{Pairing Loss}           & Train   & 0.221      & 3.72E-04         & N/A        & 0.074     & 0.196    & 0.061    \\
                                        & Test    & 1.102      & 0.934         & N/A        & 1.916     & 2.221    & 1.974    \\
\multirow{2}{*}{Pairing Accuracy}       & Train   & 94.5\%     & 100.0\%       & N/A        & 82.0\%    & 96.9\%   & 98.4\%   \\
                                        & Test    & 90.6\%     & 96.0\%        & N/A        & 58.8\%    & 88.2\%   & 87.7\%   \\ \hline
\end{tabular}
\end{center}
\end{table}

These are the representation learning prediction losses on the training and testng dataset after training for 20 epochs. The ratio between the training and testing losses can be seen in Table ~\ref{tab:ratio}.

\section{Reinforcement Learning Details}
\label{appendix:rl}

\FloatBarrier

\begin{table}[h!]
\caption{Hyperparameters for Reinforcement Learning using TRPO}
\label{tab:RLhyper}
\begin{center}

\begin{tabular}{lll}
\hline
                      & Simulation & Real Robot  \\\hline
Episode Length        & 500        & 1000        \\
Batchsize             & 2000       & 3000        \\
GAE Lamba ($\lambda$) & 0.97       & 0.97, 0.98*  \\
GAE Gamma ($\gamma$)  & 0.995      & 0.995, 0.99* \\
Max KL                & 1E-02      & 1E-02       \\
Damping Coefficient   & 1E-01      & 1E-01       \\ \hline
\end{tabular}
\end{center}
\end{table}
\FloatBarrier

These are the hyperparameters for policy learning using TRPO. We based our implementation of TRPO off of \cite{pytorchrl}. 

* On the real robot we increased $\lambda$ and decreased $\gamma$ to these values during the last hour of training as it stabilized the learning.

\end{appendices}


\end{document}